%% file: main.tex
\pdfoutput=1
\documentclass{article}
\usepackage[margin=1in]{geometry}
\usepackage{mathpazo}
\usepackage[backend=biber,style=alphabetic,natbib=true]{biblatex}
\usepackage{eqparbox}
\usepackage{multirow}

\input{header}

\title{Image Synthesis with a Single (Robust) Classifier}

\newcommand\AND{
    \end{tabular}\hfil\linebreak[4]\hfil%
    \begin{tabular}[t]{c}\ignorespaces%
}
\author{Shibani Santurkar\footnote{Equal contribution} \\
    MIT \\
  \texttt{shibani@mit.edu} \\
   \and
   Dimitris Tsipras\footnotemark[1] \\
    MIT \\
  \texttt{tsipras@mit.edu} \\
  \and 
  Brandon Tran\footnotemark[1] \hfill\null \\
    MIT \\
  \texttt{btran115@mit.edu} \\ \\
  \AND 
  Andrew Ilyas\footnotemark[1] \\
    MIT \\
  \texttt{ailyas@mit.edu} \\
  \and 
  Logan Engstrom\footnotemark[1]\\ 
    MIT \\
  \texttt{engstrom@mit.edu} \\
  \and 
  Aleksander M\k{a}dry \\
    MIT \\
  \texttt{madry@mit.edu}  
}

\date{}

\begin{document}
\maketitle
\vspace{-1em}
\begin{abstract}
\input{abstract}
\end{abstract}

\vspace{-1em}
\section{Introduction}
\input{intro}

\section{Robust Models as a Tool for Input Manipulation}
\label{sec:grad}
\input{gradients}

\section{Leveraging Robust Models for Computer Vision Tasks}
\label{sec:tasks}
\input{tasks}

\subsection{Realistic Image Generation}
\label{sec:generation}
\input{generation}

\subsection{Inpainting}
\label{sec:inpainting}
\input{inpainting}

\subsection{Image-to-Image Translation}
\label{sec:horse2zebra}
\input{horse2zebra}

\subsection{Super-Resolution}
\label{sec:superresolution}
\input{superresolution}

\subsection{Interactive Image Manipulation}
\label{sec:paint}
\input{robpaint}

\section{Discussion and Conclusions}
\input{conclusion}

\section*{Acknowledgements}
\input{acks}

\printbibliography

\clearpage

\appendix
\section{Experimental Setup}
\label{app:setup}
\input{setup}
\clearpage
\section{Omitted Figures}
\label{app:omitted}
\input{omitted}

\end{document}

%% file: header.tex
\usepackage{booktabs}
\usepackage{comment}
\usepackage[T1]{fontenc}
\usepackage[utf8]{inputenc}
\usepackage{mathtools,amssymb,amsthm}
\usepackage{bm}
\usepackage{wasysym}
\usepackage{thmtools}
\usepackage{thm-restate}
\usepackage{graphicx}
\usepackage{subcaption}
\usepackage{floatrow}
\usepackage[]{todonotes}
\usepackage{textgreek}

\usepackage[pdfencoding=auto]{hyperref}

\newcolumntype{P}[1]{>{\centering\arraybackslash}p{#1}}

\DeclareMathOperator*{\argmin}{arg\,min}
\DeclareMathOperator*{\argmax}{arg\,max}

\newcommand{\HtoZ}{Horse $\leftrightarrow$ Zebra}
\newcommand{\StoW}{Summer $\leftrightarrow$ Winter}
\newcommand{\AtoO}{Apple $\leftrightarrow$ Orange}

\newcommand{\D}{\mathcal{D}}
\newcommand{\Gy}{\mathcal{G}_y}
\newcommand{\loss}{\mathcal{L}}

\newcommand{\eps}{\varepsilon}
\newcommand{\E}{\mathbb{E}}

\let\oldmu\mu
\let\oldsig\Sigma
\renewcommand{\mu}{\bm{\oldmu}}
\renewcommand{\Sigma}{\bm{\oldsig}}

\newfloatcommand{capbtabbox}{table}[][\FBwidth]

%% file: abstract.tex
We show that the basic classification framework alone can be used to
tackle some of the most challenging tasks in image synthesis.
In contrast to other state-of-the-art approaches, the
toolkit we develop is rather minimal: it uses a {\em single, off-the-shelf}
classifier for {\em all} these tasks.
The crux of our approach is that we train this classifier to be {\em adversarially robust}.
It turns out that adversarial robustness is precisely what we need
to directly manipulate salient features of the input.
Overall, our findings demonstrate the utility of robustness in the broader
machine learning context.\footnote{Code and models for our experiments can be
found at \url{https://git.io/robust-apps}.}

%% file: intro.tex
Deep learning has revolutionized the way we tackle computer vision problems.
This revolution started with progress on image
classification~\cite{krizhevsky2012imagenet,he2015delving,he2016deep},
which then triggered the expansion of the deep learning paradigm to encompass
more sophisticated tasks such as
image generation~\cite{karras2018progressive,brock2019large}
and image-to-image translation~\cite{isola2017image,zhu2017unpaired}.
Much of this expansion was predicated on developing complex,
task-specific techniques, often rooted in
the generative adversarial network (GAN) framework~\cite{goodfellow2014generative}.
However, {\em is there a simpler toolkit for solving these tasks}?

\begin{figure}[!h]
	\centering
	\includegraphics[width=1\textwidth]{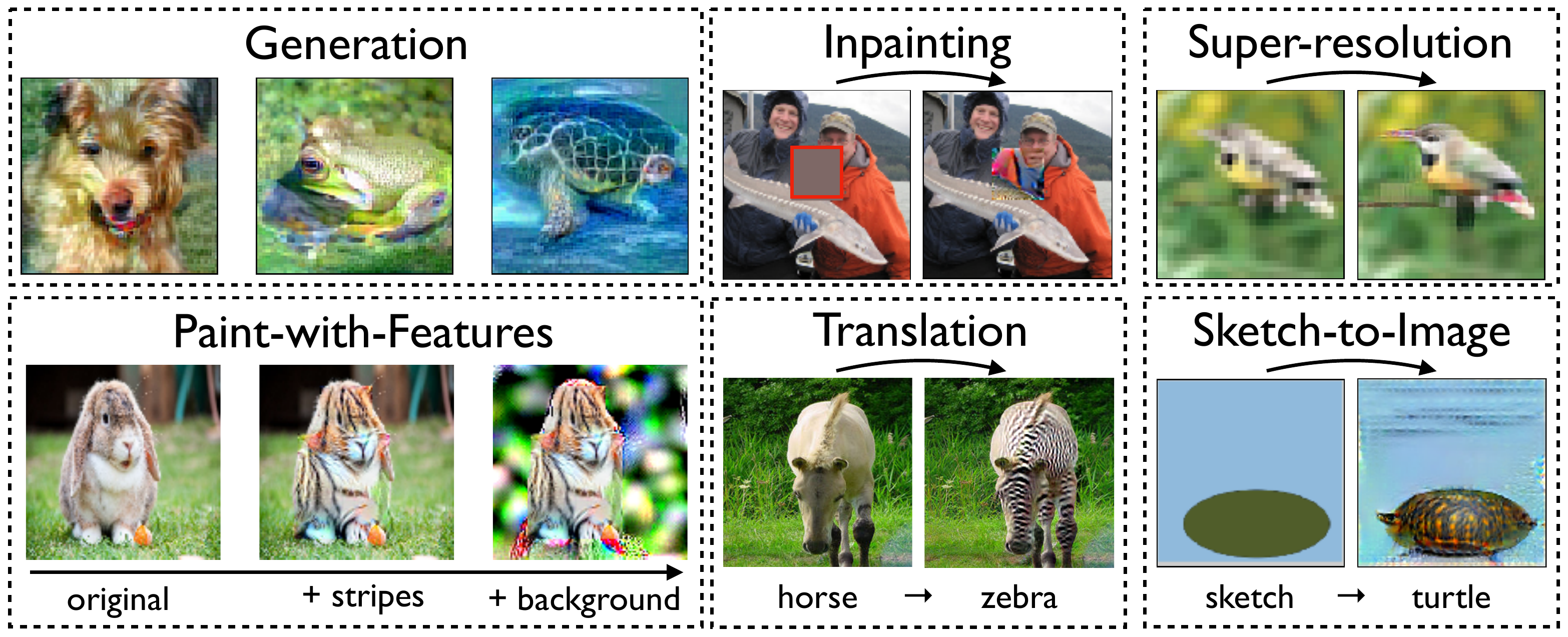}
	\caption{Image synthesis and manipulation tasks performed using a \emph{single} 
					(robustly trained) classifier.}
	\label{fig:intro_fig}
\end{figure}

In this work, we demonstrate that basic classification tools {\em alone}
suffice to tackle various image synthesis tasks. These tasks include
(cf. Figure~\ref{fig:intro_fig}):
{\em generation} (Section~\ref{sec:generation}),
{\em inpainting} (Section~\ref{sec:inpainting}),
{\em image-to-image translation} (Section~\ref{sec:horse2zebra}),
{\em super-resolution} (Section~\ref{sec:superresolution}), and
{\em interactive image manipulation} (Section~\ref{sec:paint}).

Our entire toolkit is based on {\em a single classifier} (per dataset)
and involves performing a simple input
manipulation: maximizing predicted class scores with gradient descent.
Our approach is thus general purpose and simple to implement and train, 
while also requiring minimal tuning.
To highlight the potential of the core methodology itself, 
we intentionally employ a generic classification setup
(ResNet-50~\cite{he2016deep} with default hyperparameters) without any
additional optimizations (e.g., domain-specific priors or regularizers).
Moreover, to emphasize the consistency of our approach,
throughout this work we demonstrate performance on {\em randomly selected}
examples from the test set.

The key ingredient of our method is {\em
adversarially robust} classifiers.
Previously, Tsipras et al.~\cite{tsipras2019robustness} observed that maximizing
the loss of  robust models over the input leads to realistic
instances of other classes.
Here we are able to fully leverage this connection to build a versatile toolkit
for image synthesis.
Our findings thus establish robust classifiers as a powerful primitive
for semantic image manipulation, despite them being trained {\em solely} to perform image 
classification.

%% file: gradients.tex
Recently, Tsipras et al.~\cite{tsipras2019robustness} observed 
that optimizing an image to cause a misclassification in an (adversarially)
robust classifier introduces salient characteristics of the 
incorrect class.  This property is unique to {\em robust} classifiers: 
standard models (trained with empirical risk minimization (ERM)) are 
inherently brittle, and their predictions are sensitive even to 
imperceptible changes in the input~\cite{szegedy2014intriguing}.

Adversarially robust classifiers are trained using the {\em robust 
optimization} objective~\cite{wald1945statistical,madry2018towards}, where instead of
minimizing the expected loss $\loss$ over the data
\begin{equation}
\E_{(x,y)\sim\D} \left[\loss(x,y)\right],
\end{equation}
we minimize the worst case loss over a specific perturbation 
set $\Delta$
\begin{equation}
\E_{(x,y)\sim\D} \left[\max_{\delta\in\Delta}\ \loss(x+\delta,y)\right].
\label{eq:ro}
\end{equation}
Typically, the set $\Delta$ captures
imperceptible changes (e.g., small $\ell_2$ perturbations), 
and given such a $\Delta$, the problem in~\eqref{eq:ro} can be solved using adversarial 
training~\cite{goodfellow2015explaining,madry2018towards}.

From one perspective, we can view robust optimization as encoding priors into
the model, preventing it from relying on imperceptible features of the input~\cite{engstrom2019learning}.
Indeed, the findings of Tsipras et al.~\cite{tsipras2019robustness} are aligned
with this viewpoint---by encouraging the model to be invariant to small
perturbations, robust training ensures that changes in the model's predictions
correspond to salient input changes.

\begin{figure}[!h]
	\includegraphics[width=\textwidth]{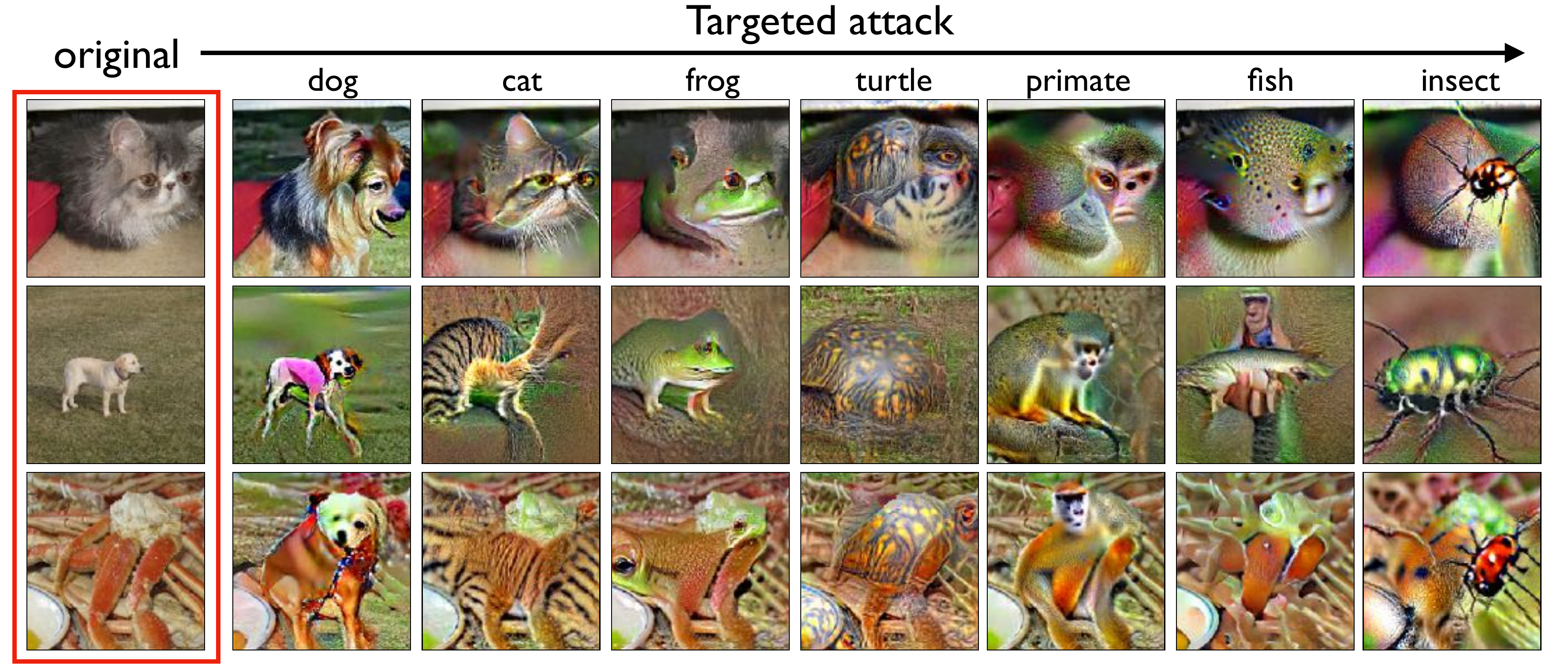}
	\caption{Maximizing class scores of a robustly trained classifier.
		For each original image, we visualize the result of performing targeted 
        projected gradient descent (PGD) toward different classes. The resulting
    images actually resemble samples of the target class.}
	\label{fig:targeted}
\end{figure}

In fact, it turns out that this phenomenon also emerges when we maximize
the probability of a {\em specific class} (targeted attacks)
for a robust model---see Figure~\ref{fig:targeted} for an illustration.
This indicates that robust models exhibit more human-aligned
gradients, and, more importantly, that we can precisely control features
in the input just by performing gradient descent on the model output.
Previously, performing such manipulations has only been possible with more
complex and task-specific
techniques~\cite{mordvintsev2015inceptionism,radford2016unsupervised,isola2017image,zhu2017unpaired}. In the
rest of this work, we demonstrate that this property of robust models is
sufficient to attain good performance on a diverse set of image synthesis tasks.

%% file: tasks.tex
Deep learning-based methods have recently made significant progress on
image synthesis and manipulation tasks, typically by training
specifically-crafted models in the {GAN}
framework~\cite{goodfellow2014generative,iizuka2017globally,
	zhu2017unpaired,yu2018generative,brock2019large}, using priors obtained from deep generative
models~\cite{nguyen2016synthesizing,nguyen2017plug,ulyanov2017deep,yeh2017semantic}, or leveraging standard
classifiers via sophisticated, task-specific
methods~\cite{mordvintsev2015inceptionism,oygard2015visualizing,tyka2016class,gatys2016image}.
We discuss additional related work in the following subsections as necessary.

In this section, we outline our methods and results for obtaining competitive
performance on these tasks using only robust (feed-forward) classifiers. Our
approach is remarkably simple: all the applications are performed using
gradient ascent on class scores derived from the same robustly
trained classifier. In particular, it does not involve
fine-grained tuning (see Appendix~\ref{app:tune}), highlighting the potential of
robust classifiers as a versatile primitive for sophisticated vision tasks.

%% file: generation.tex
Synthesizing realistic samples for natural data domains (such as 
images) has been a long standing challenge in computer vision.
Given a set of example inputs, we would like to 
learn a model that can produce novel perceptually-plausible inputs. The 
development of deep learning-based methods such as autoregressive
models~\cite{hochreiter1997long,graves2013generating,van2016pixel},
auto-encoders~\cite{vincent2010stacked,kingma2013autoencoding} and flow-based
models~\cite{dinh2014nice,rezende2015variational,dinh2017density,kingma2018glow}
has led to significant progress in this domain.
More recently, advancements in 
generative adversarial networks (GANs)~\cite{goodfellow2014generative} 
have made it possible to generate high-quality images for challenging
datasets~\cite{zhang2018self,karras2018progressive,brock2019large}.
Many of these methods, however, can be tricky to train and properly 
tune. They are also fairly computationally intensive, and often require fine-grained 
performance optimizations.

In contrast, we demonstrate that robust classifiers, without any special 
training or auxiliary networks, can be a powerful tool for synthesizing 
realistic natural images. At a high level, our generation procedure 
is based on maximizing the class score of the desired class using a 
robust model. The purpose of this maximization is to add relevant and semantically 
meaningful features of that class to a given input image. 
This approach has been previously used on standard models to perform class
visualization---synthesizing prototypical inputs of each class---in combination
with domain-specific input priors (either hand-crafted~\cite{nguyen2015deep} and
learned~\cite{nguyen2016synthesizing,nguyen2017plug}) or
regularizers~\cite{simonyan2013deep,mordvintsev2015inceptionism,oygard2015visualizing,tyka2016class}.

As the process of class score maximization is deterministic, generating a diverse set of samples requires a random
seed as the starting point of the maximization process. 
Formally, to generate a sample of class $y$, we sample a seed and minimize the
loss $\loss$ of label $y$
$$x = \argmin_{\|x' - x_0\|_2\leq\eps}\loss(x', y), \qquad x_0\sim \Gy,$$
 for some class-conditional seed distribution $\Gy$, using projected gradient descent (PGD)
(experimental details can be found in Appendix~\ref{app:setup}). 
Ideally, samples from $\Gy$ should be diverse and  
statistically similar to the data distribution. Here, we use 
a simple (but already sufficient) choice for $\Gy$---a multivariate normal
distribution fit to the empirical class-conditional distribution
$$\Gy := \mathcal{N}(\mu_y, \Sigma_y),\quad \text{ where }
    \mu_y = \E_{x\sim\D_y}[x], \; \Sigma = \E_{x\sim\D_y}[(x-\mu_y)^\top (x
    -\mu_y)],$$
and $\D_y$ is the distribution of natural inputs conditioned on the label
$y$.
We visualize example seeds from these multivariate Gaussians in
Figure~\ref{fig:seeds}.
\begin{figure}[!h]
    \begin{subfigure}[b]{1\textwidth}
	\includegraphics[width=\textwidth]{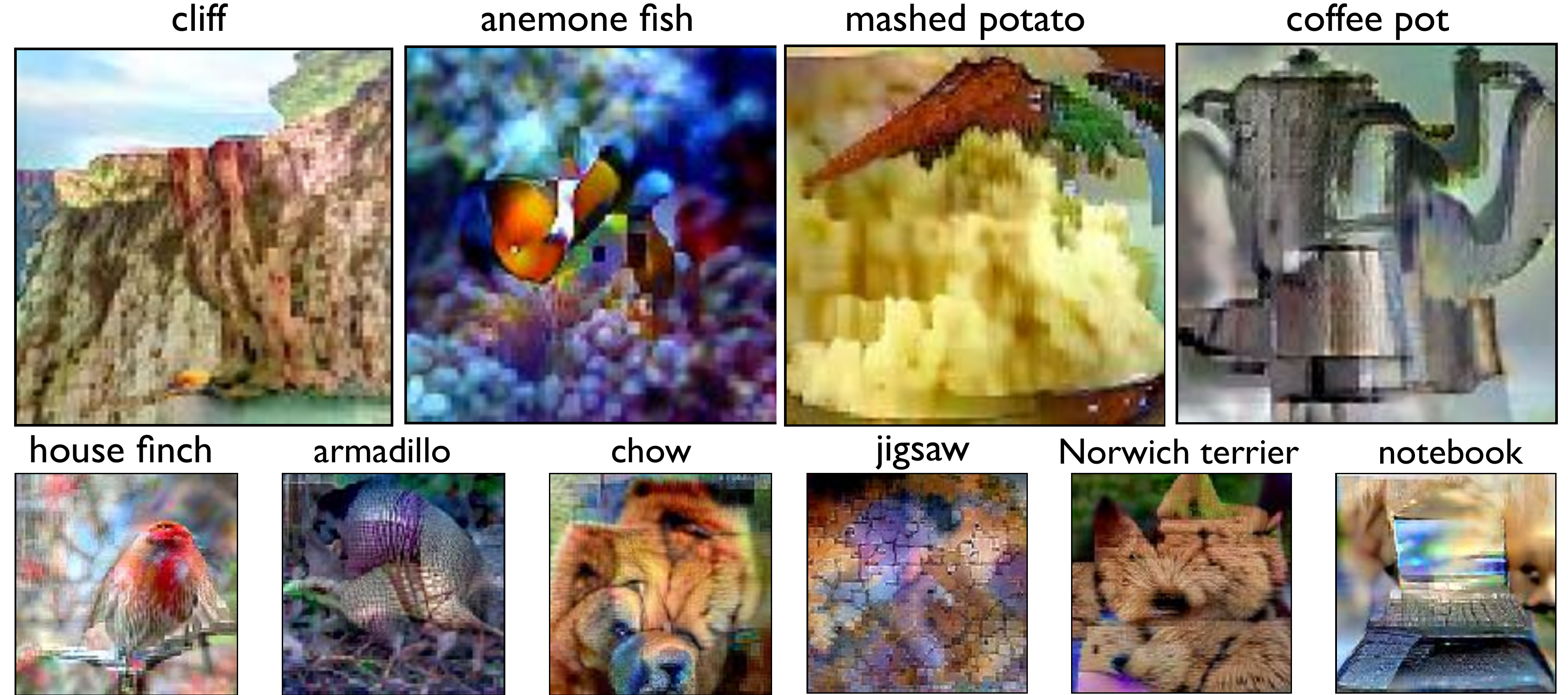}
    \caption{}
    \end{subfigure}
    \begin{subfigure}[b]{1\textwidth}
	\includegraphics[width=\textwidth]{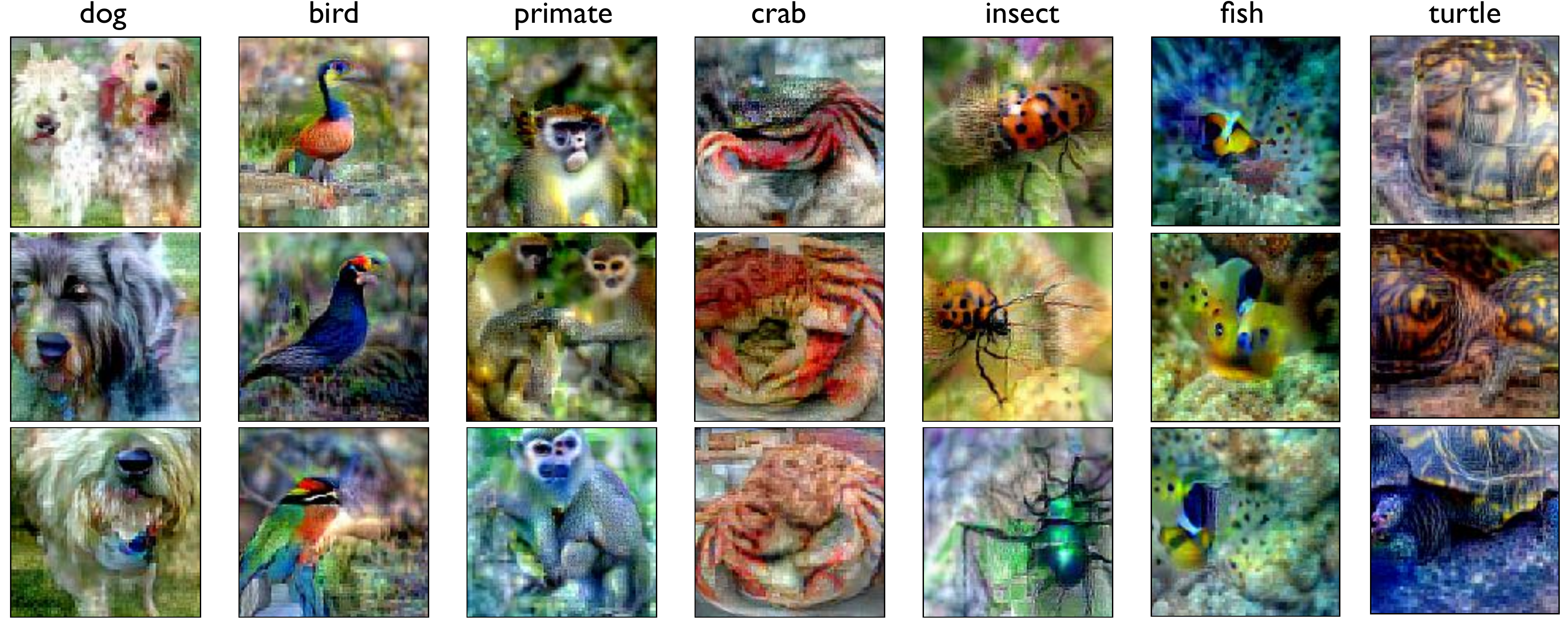}
    \caption{}
	\end{subfigure}
    \caption{{\em Random} samples (of resolution 224$\times$224) produced using 
        a robustly trained classifier. We show: (a) samples from several (random) 
        classes of the ImageNet dataset and (b) multiple samples from a few
        random classes of the restricted ImageNet dataset (to illustrate
        diversity). See Figures~\ref{fig:CIFAR_gen_full},
        \ref{fig:restricted_gen_full}, \ref{fig:imagenet_gen_full}, 
        and \ref{fig:diversity} of Appendix~\ref{app:omitted} for additional 
        samples.}
	\label{fig:generation}
\end{figure}

This approach enables us to perform
\emph{conditional} image synthesis given any target class.
Samples (at resolution 224$\times$224) produced by our method are shown in 
Figure~\ref{fig:generation} (also see Appendix~\ref{app:omitted}).
The resulting images are diverse
and  realistic, despite the fact that they are
generated using targeted PGD on off-the-shelf
robust models without any additional optimizations.
\footnote{Interestingly, the robust model used to generate
these high-quality ImageNet samples is only $45\%$ 
accurate, yet has a sufficiently rich representation to synthesize 
semantic features for $1000$ classes.}

\paragraph{Different seed distributions.} It is worth noting that there is significant 
room for improvement in designing the distribution $\Gy$. One way 
to synthesize better samples would be to use a richer distribution---for 
instance, mixtures of Gaussians per class to better capture multiple data modes. 
Also, in contrast to many existing approaches, we are not limited to a single seed distribution,
and we could even utilize other methods (such as procedural generation) 
to customize seeds with specific structure or color, and then maximize 
class scores to produce realistic samples (e.g., see Section~\ref{sec:paint}).

\paragraph{Evaluating Sample Quality.}
Inception Score (IS)~\cite{salimans2016improved} is a popular metric for
evaluating the quality of generated image data. 
Table~\ref{tab:inception_score} presents the IS of samples
generated using a robust classifier.

{\renewcommand{\arraystretch}{1}%
\begin{table}[!h]
	\caption{Inception Scores (IS) for samples generated using robustly trained
	classifiers compared to state-of-the-art 
	generation approaches~\cite{gulrajani2017improved,shmelkov2018good, brock2019large} (cf. Appendix~\ref{app:is} 
	for details).}
\begin{tabular}{P{2.3cm}|P{2.3cm}P{2.3cm}P{2.35cm}P{2.6cm} }
	\toprule
    Dataset & Train Data & BigGAN~\cite{brock2019large} &
        WGAN-GP~\cite{gulrajani2017improved} & {Our approach} \\
	\midrule
	CIFAR-10   & 11.2 $\pm$ 0.2    & 9.22 &    8.4 $\pm$ 0.1 & {7.5 $\pm$ 0.1} \\
	ImageNet\footnotemark &   331.9 $\pm$ 4.9  & 233.1 $\pm$ 1    & 11.6 & {259.0 $\pm$ 4} \\
	\bottomrule
\end{tabular}
\label{tab:inception_score}
\end{table}
\footnotetext[1]{For ImageNet, there is a difference in resolution between BigGAN samples 
	($256\times256$), SAGAN ($128\times128$) and our approach ($224\times 224$). BigGAN 
	attains IS of 166.5.
	at $128\times 128$ resolution.
}

We find that our approach improves over state-of-the-art
(BigGAN~\cite{brock2019large}) in
terms of Inception Score on the ImageNet dataset, yet, at the same time, the
Fr\'echet Inception Distance (FID)~\cite{heusel2017gans} is worse
(36.0 versus 7.4). These results can be explained by the fact that, on one hand, our samples are essentially adversarial examples (which are
known to transfer across models~\cite{szegedy2014intriguing}) and thus are likely
to induce highly confident predictions that IS is designed to pick up. On the other hand, GANs are explicitly trained to produce samples that are indistinguishable from true data with respect to a discriminator, and hence are likely to have a better (lower) FID.

%


%% file: inpainting.tex
Image inpainting is the task of recovering images with large 
corrupted regions~\cite{efros1999texture,bertalmio2000image,hays2007scene}.
Given an image $x$, corrupted in a region 
corresponding to a binary mask $m \in \{0, 1\}^d$, the goal 
of inpainting is to recover the missing pixels in a manner that is perceptually
plausible with respect to the rest of the image.
We find that simple feed-forward classifiers,
when robustly trained, can be a powerful tool for such image 
reconstruction tasks. 

From our perspective, the goal is to use robust models to restore missing features
of the image. To this end, we will optimize the image to 
maximize the score of the underlying true class, while also 
forcing it to be consistent with the original in the uncorrupted 
regions. Concretely, given a robust classifier trained on 
uncorrupted data, and a corrupted image $x$
with label $y$, we solve
\begin{equation}
x_I = \argmin_{x'} \mathcal{L}(x', y) + \lambda || (x - x') \odot (1 - m)||_2
\label{eq:inpaint}
\end{equation}
where $\mathcal{L}$
is the cross-entropy loss, $\odot$ denotes element-wise multiplication, and $\lambda$ is an appropriately chosen constant.
Note that while we require knowing the underlying label $y$ for the input, it can typically be
accurately predicted by the classifier itself given the corrupted image.

\begin{figure}[!h]
	\centering
	\begin{subfigure}[b]{0.49\textwidth}
	\centering
	\includegraphics[width=0.98\textwidth]{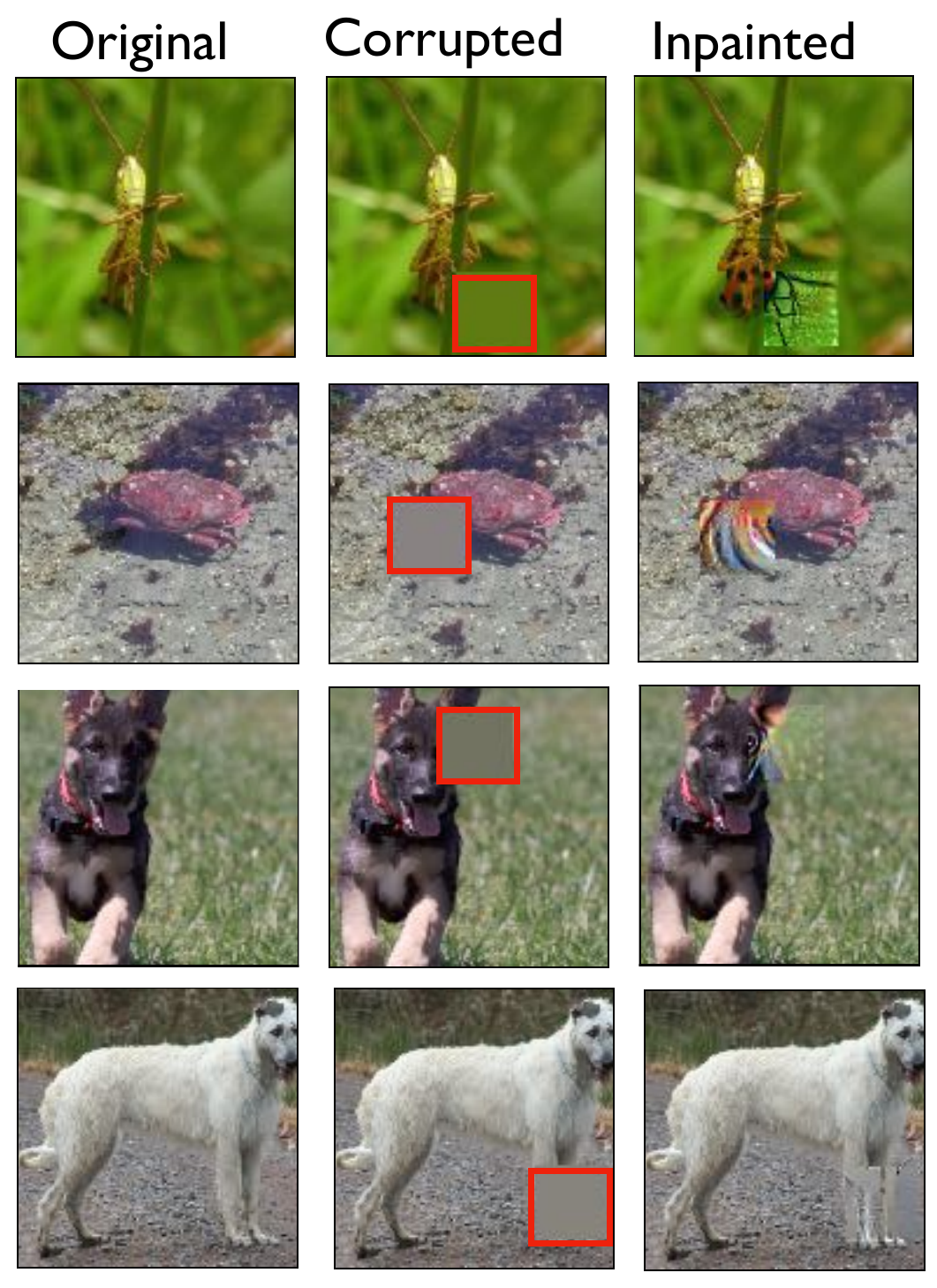}
	\caption{\emph{random} samples}
	\label{fig:inpaint_rand}
	\end{subfigure}
\hfil
	\begin{subfigure}[b]{0.49\textwidth}
	\centering
	\includegraphics[width=.99\textwidth]{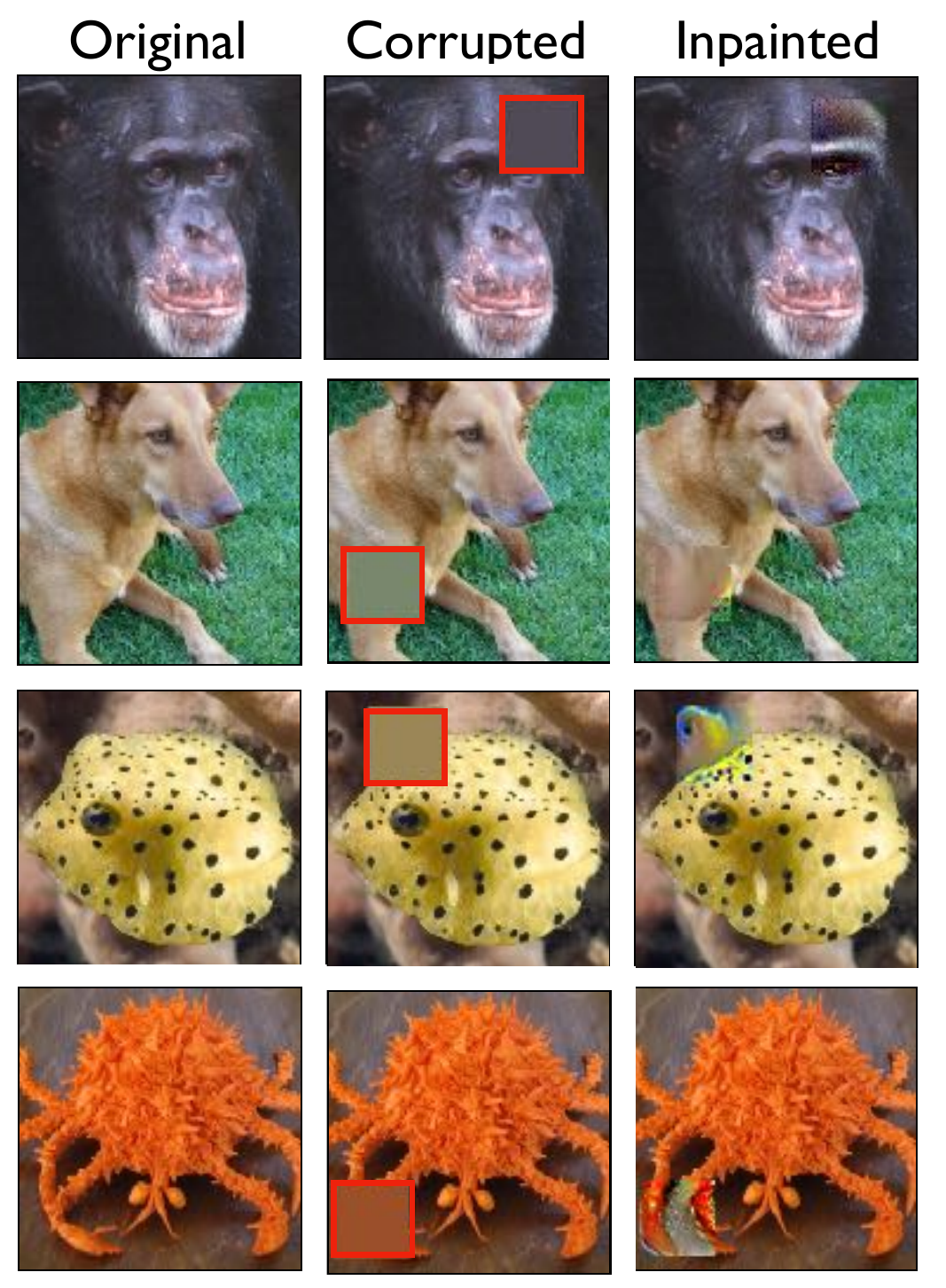}
	\caption{select samples}
	\label{fig:inpaint_select}
\end{subfigure}
	\caption{Image inpainting using robust models -- \textit{left:} original,
	\textit{middle:} corrupted and \textit{right:} inpainted samples. 
	To recover 
	missing regions, we use PGD to maximize the class score predicted for the 
	image while penalizing changes to 
    the uncorrupted regions. }
	\label{fig:inpainting}
\end{figure}

In Figure~\ref{fig:inpainting}, we show sample reconstructions obtained by
optimizing~\eqref{eq:inpaint} using PGD (cf. Appendix
~\ref{app:setup} for details). We can observe that these reconstructions
look remarkably similar to the uncorrupted images in terms of semantic 
content. Interestingly, even when this approach fails (reconstructions
differ from the original), the resulting images do tend to be perceptually
plausible to a human, as shown in Appendix Figure~\ref{fig:inpainting_bloopers}.

%% file: horse2zebra.tex
As discussed in Section~\ref{sec:grad}, robust 
models provide a mechanism for transforming inputs between classes.
In computer vision literature, this would be an instance of
{\em image-to-image translation}, where the goal is to translate an image
from a source to a target
domain in a semantic manner~\cite{hertzmann2001image}.

In this section, we demonstrate that robust classifiers give rise to a new methodology
for performing such image-to-image translations.
The key is  to (robustly) train a classifier to distinguish between
the source and target domain.
Conceptually, such a classifier will extract salient characteristics of each
domain in order to make accurate predictions.
We can then translate an input from the source domain by
directly maximizing the predicted score of the target domain.

\begin{figure}[htp]
    \begin{center}
        \begin{tikzpicture}[scale=2]
        \node at (7.1,0) {
            \includegraphics[width=0.13\textwidth,trim={.3cm 0 0 0},clip]{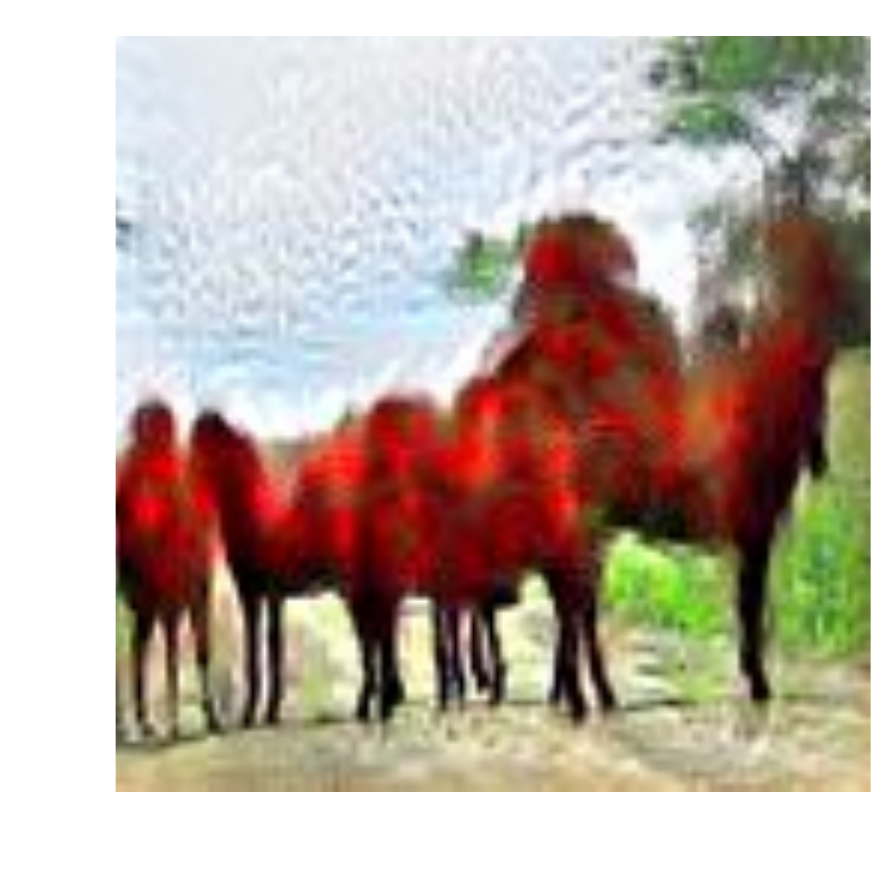}};
        \node at (6.1,0) {
            \includegraphics[width=0.13\textwidth,trim={.3cm 0 0 0}, clip]{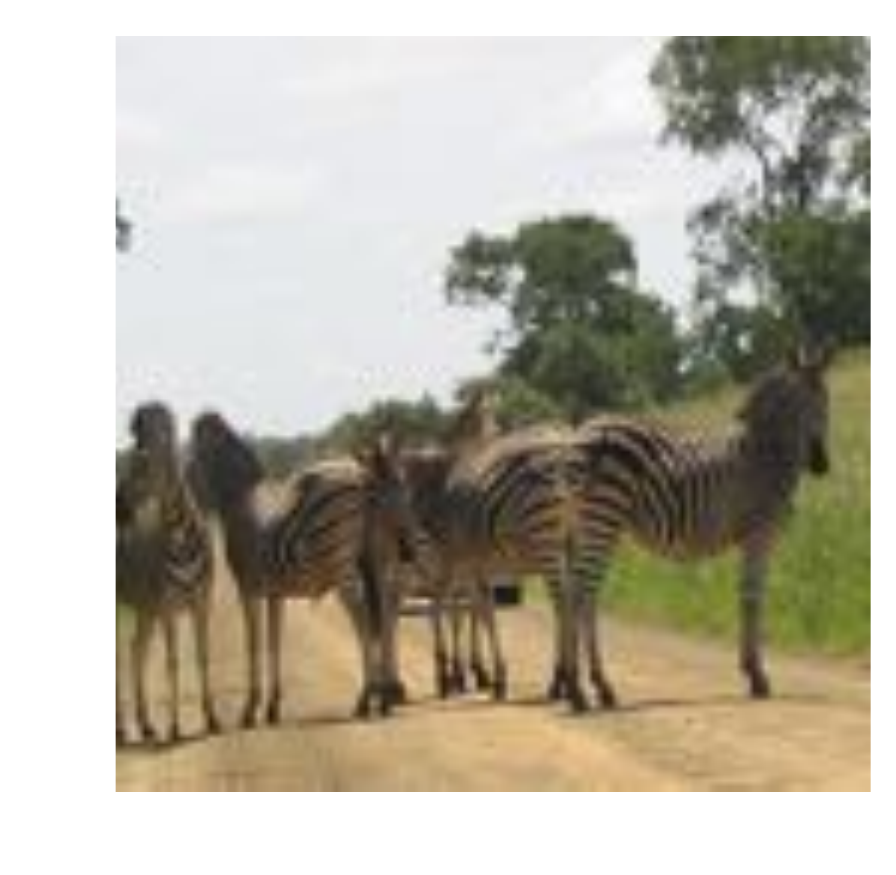}};
        \node at (5.1,0) {
            \includegraphics[width=0.13\textwidth,trim={.3cm 0 0 0}, clip]{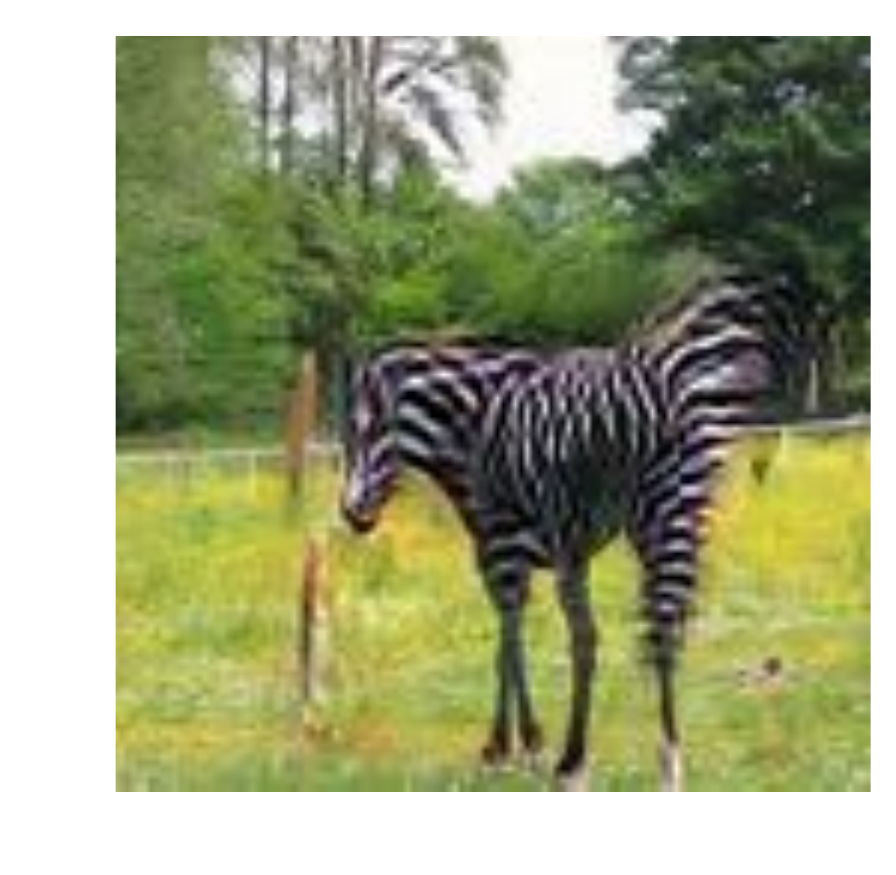}};
        \node at (4.1,0) {
            \includegraphics[width=0.13\textwidth,trim={.3cm 0 0 0}, clip]{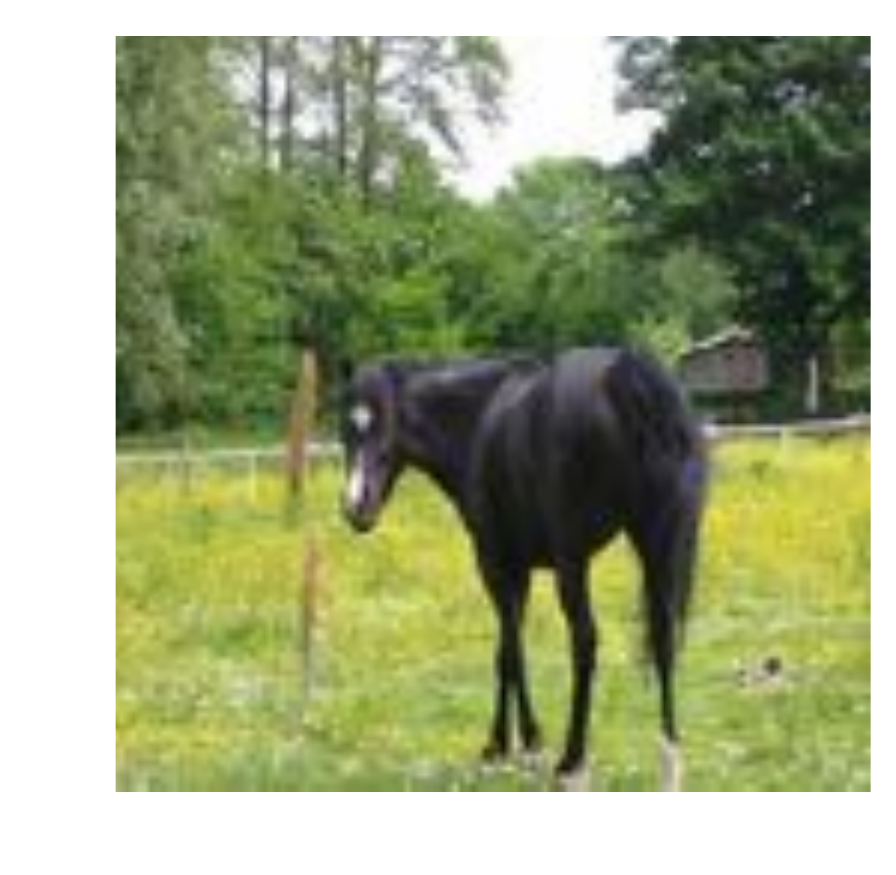}};
        \node at (3,0) {
            \includegraphics[width=0.13\textwidth,trim={.3cm 0 0 0}, clip]{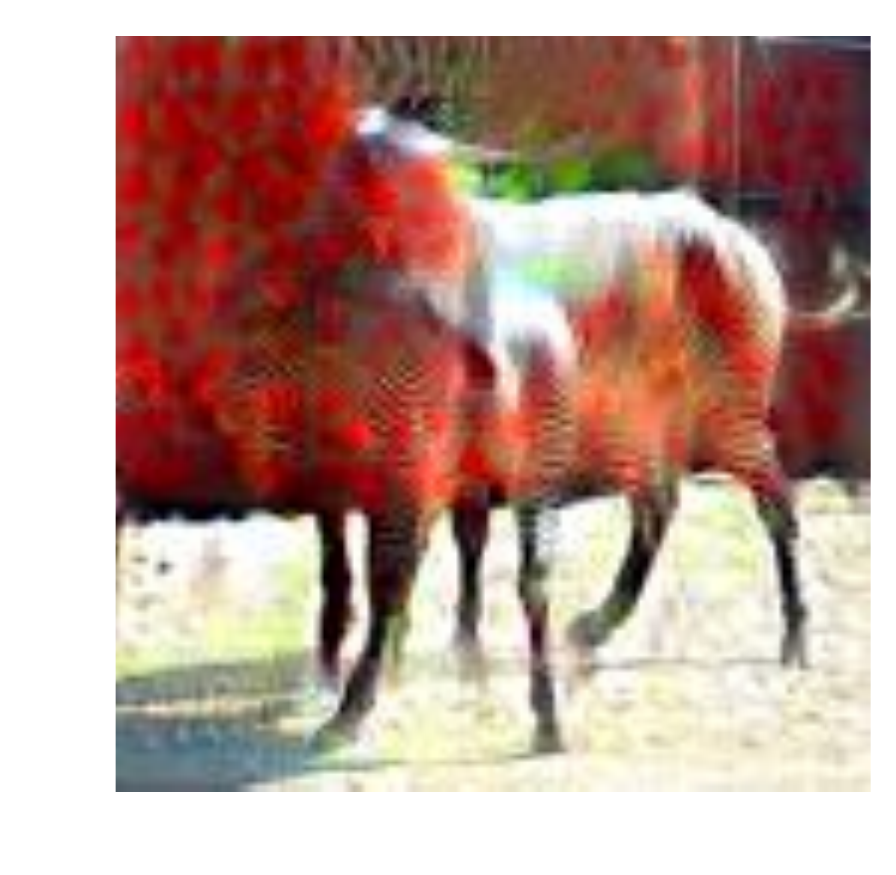}};
        \node at (2,0) {
            \includegraphics[width=0.13\textwidth,trim={.3cm 0 0 0}, clip]{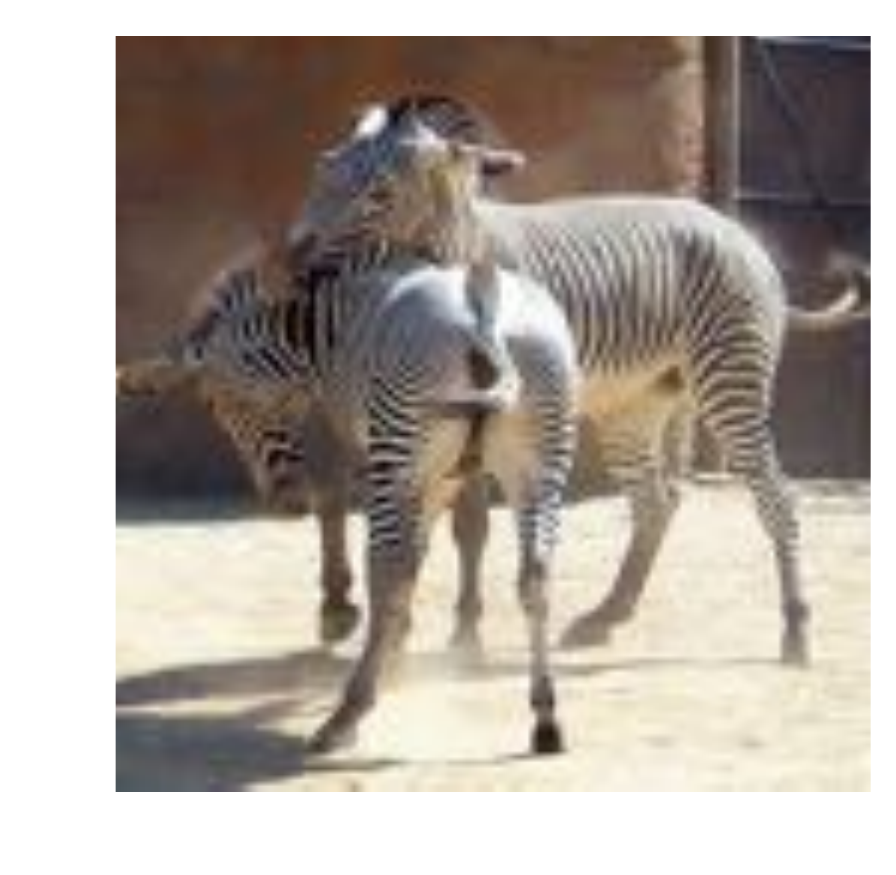}};
        \node at (1,0) {
            \includegraphics[width=0.13\textwidth,trim={.3cm 0 0 0}, clip]{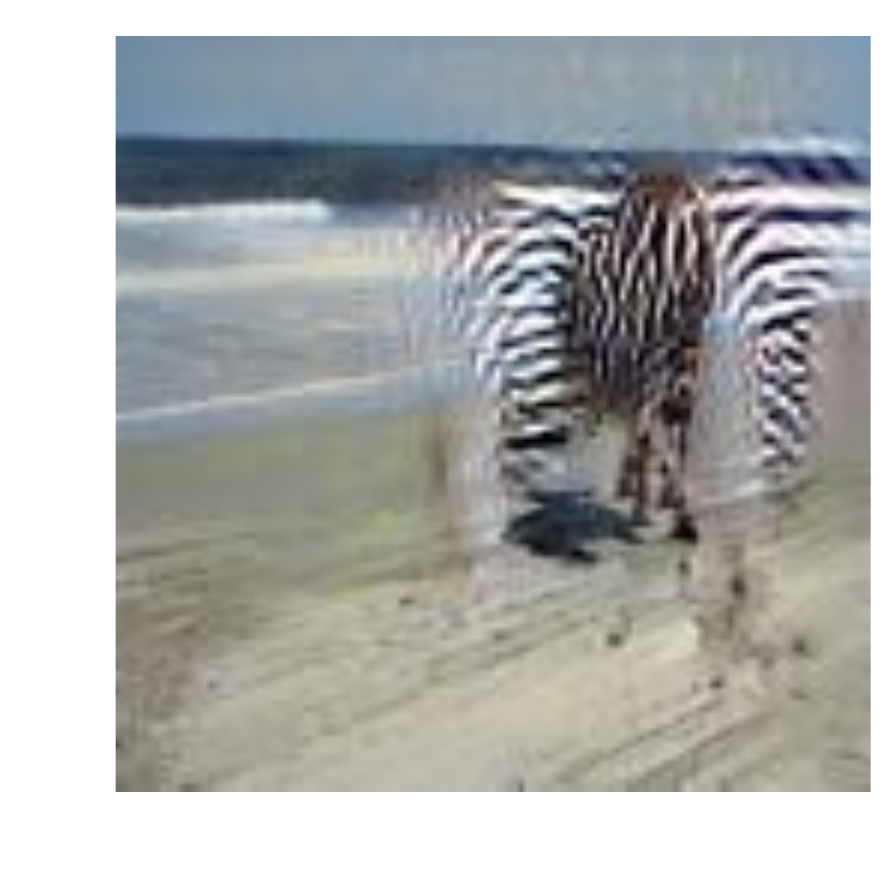}};
        \node at (0,0) {
            \includegraphics[width=0.13\textwidth,trim={.3cm 0 0 0}, clip]{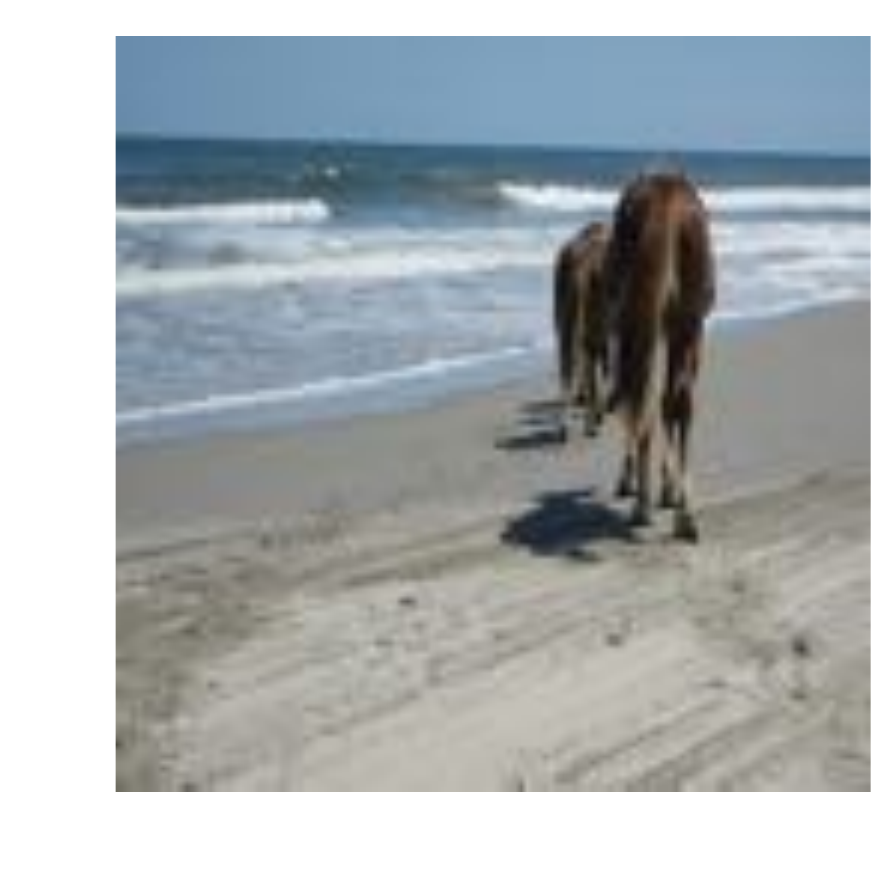}};

        \node at (0.5, 0.65) {horse $\to$ zebra};
        \node at (2.5, 0.65) {zebra $\to$ horse};
        \node at (4.55, 0.65) {horse $\to$ zebra};
        \node at (6.55, 0.65) {zebra $\to$ horse};

        \node at (7.1,-1.3) {
            \includegraphics[width=0.13\textwidth,trim={.3cm 0 0 0}, clip]{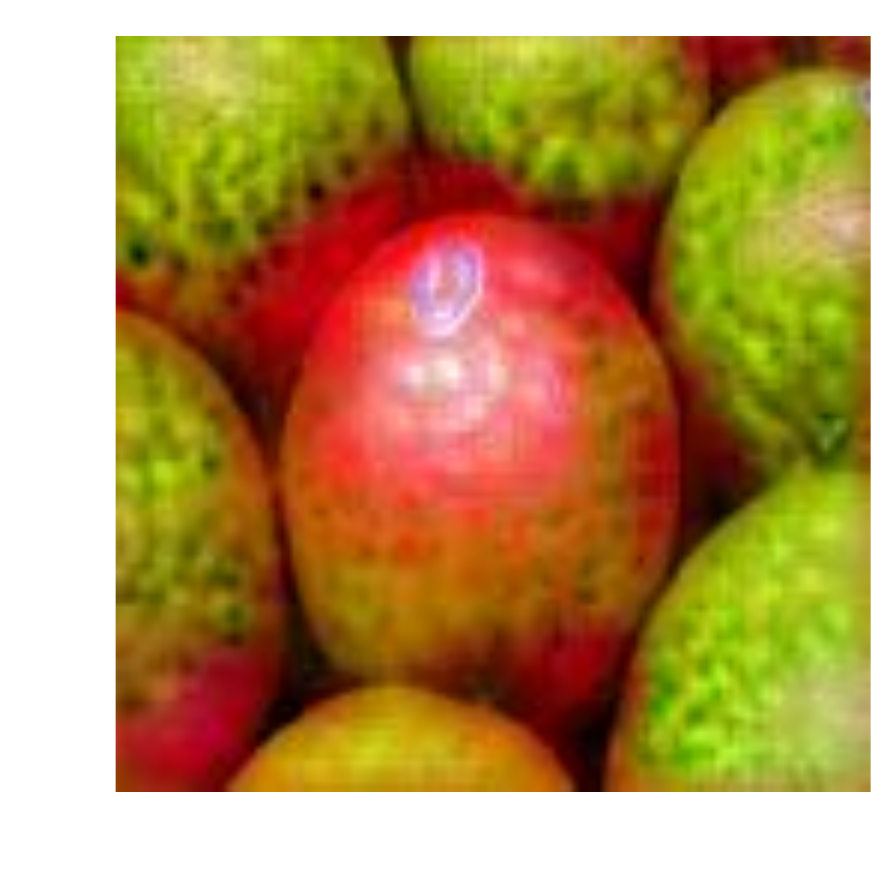}};
        \node at (6.1,-1.3) {
            \includegraphics[width=0.13\textwidth,trim={.3cm 0 0 0}, clip]{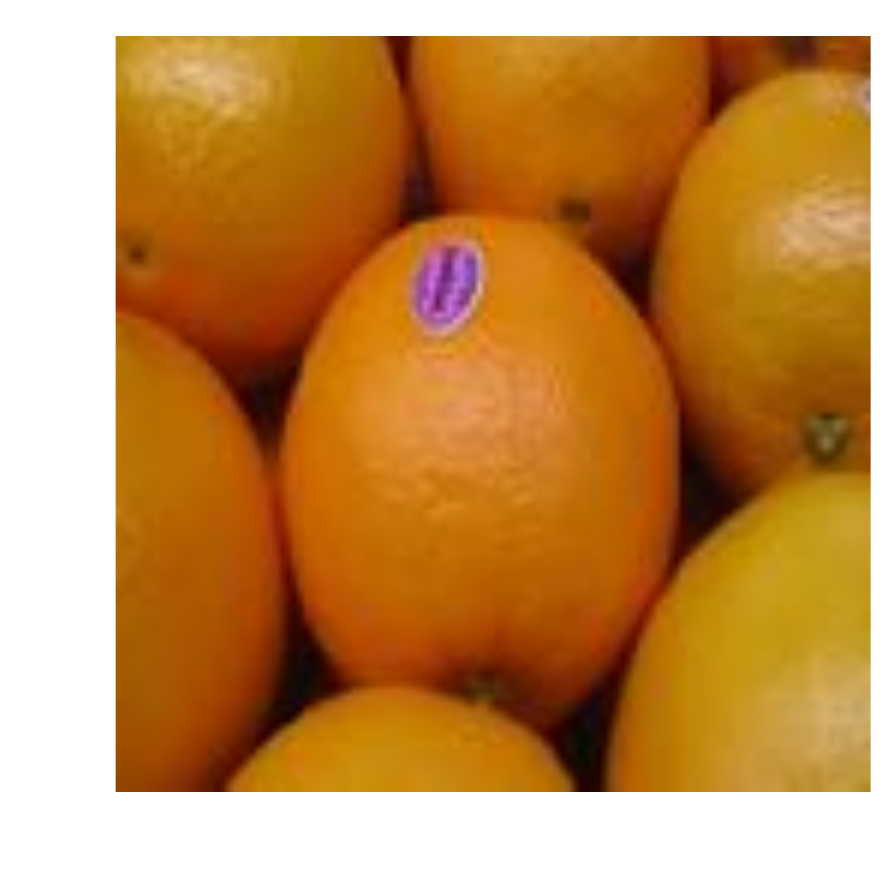}};
        \node at (5.1,-1.3) {
            \includegraphics[width=0.13\textwidth,trim={.3cm 0 0 0}, clip]{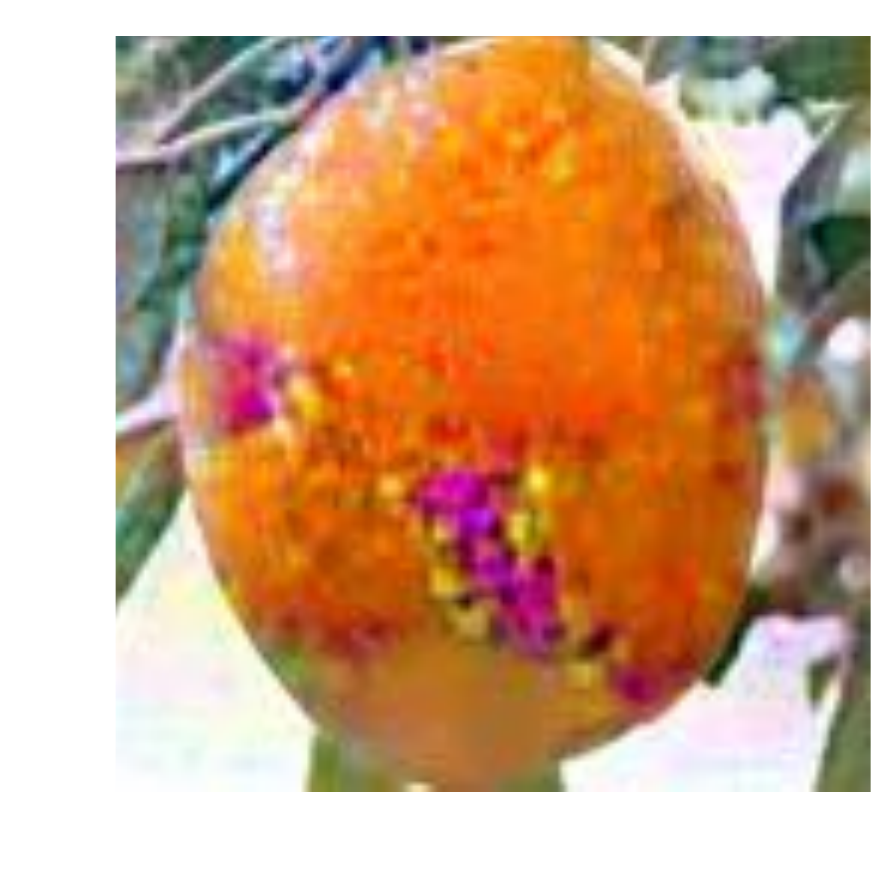}};
        \node at (4.1,-1.3) {
            \includegraphics[width=0.13\textwidth,trim={.3cm 0 0 0}, clip]{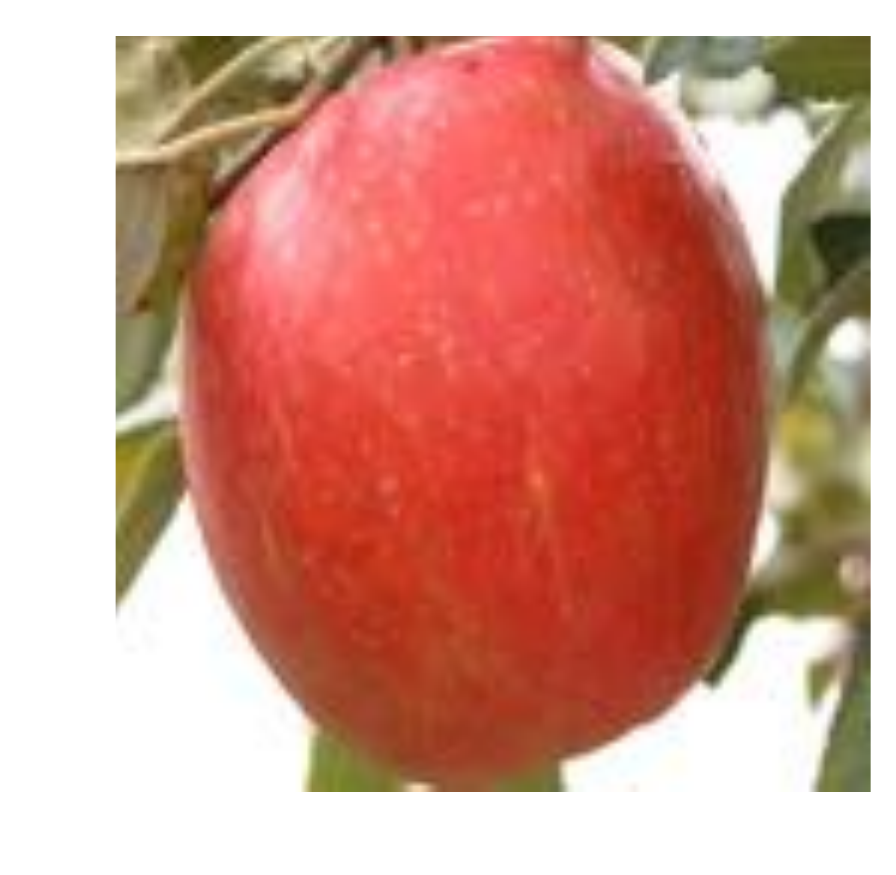}};
        \node at (3,-1.3) {
            \includegraphics[width=0.13\textwidth,trim={.3cm 0 0 0}, clip]{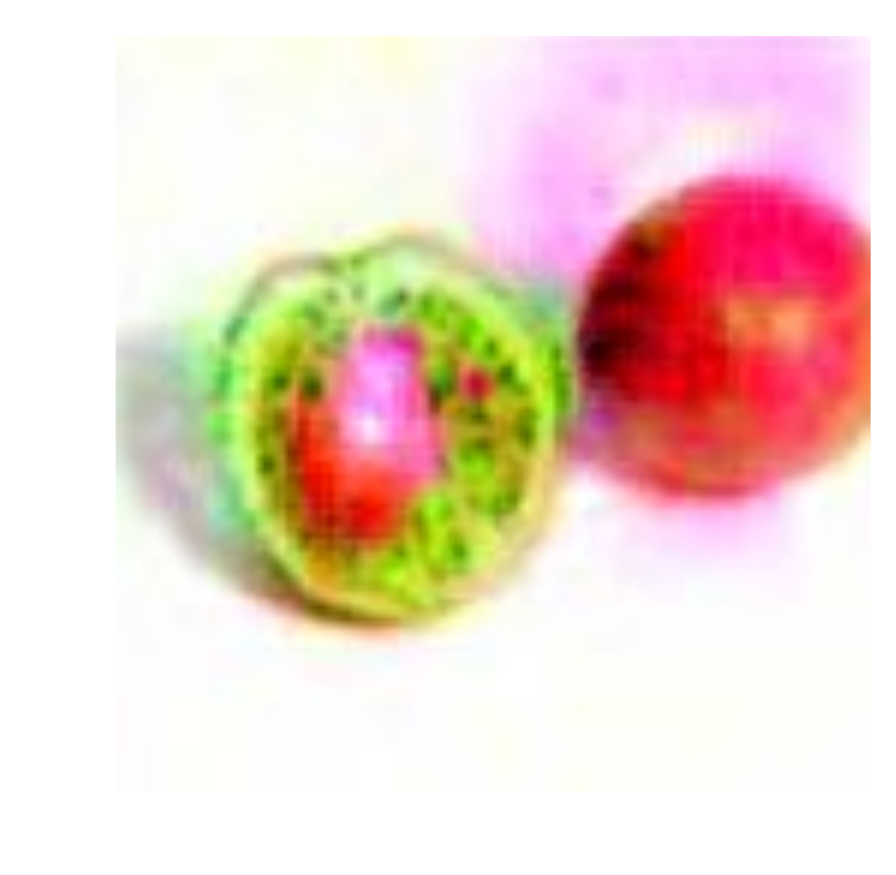}};
        \node at (2,-1.3) {
            \includegraphics[width=0.13\textwidth,trim={.3cm 0 0 0}, clip]{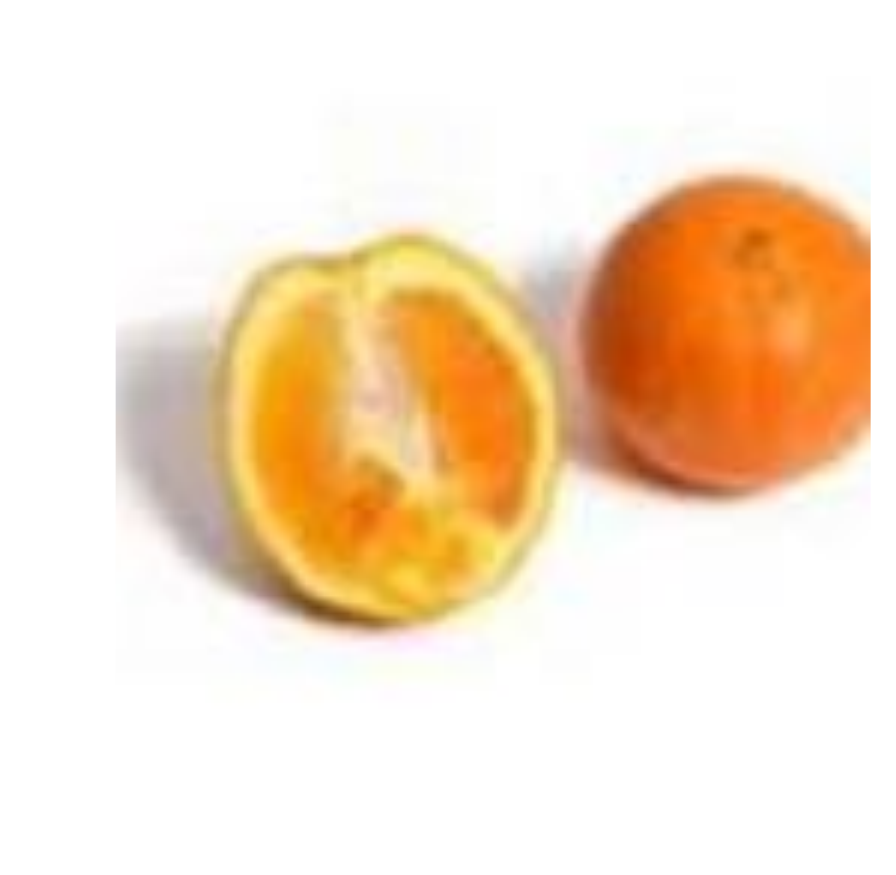}};
        \node at (1,-1.3) {
            \includegraphics[width=0.13\textwidth,trim={.3cm 0 0 0}, clip]{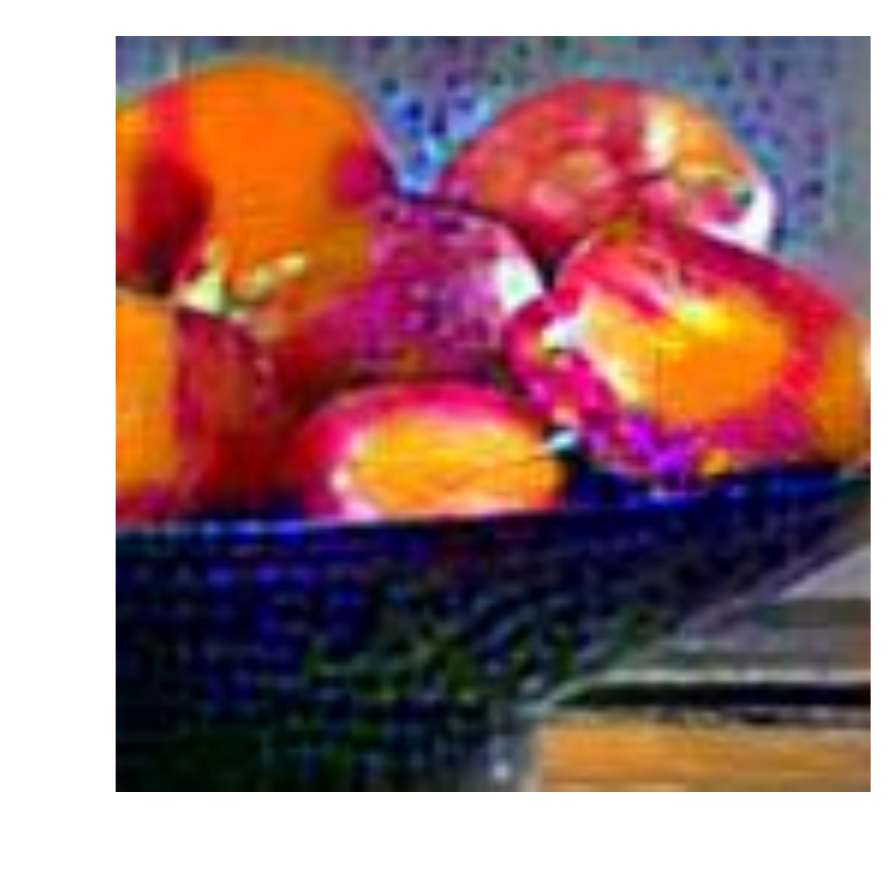}};
        \node at (0,-1.3) {
            \includegraphics[width=0.13\textwidth,trim={.3cm 0 0 0}, clip]{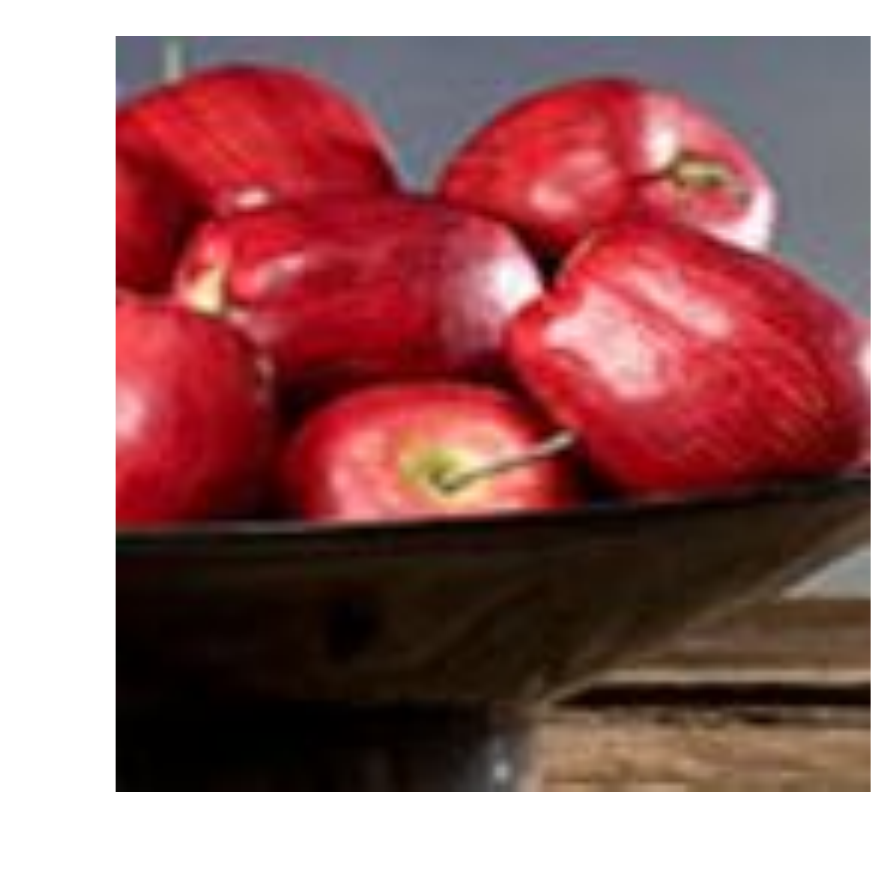}};

        \node at (0.5, -0.65) {apple $\to$ orange};
        \node at (2.5, -0.65) {orange $\to$ apple};
        \node at (4.55, -0.65) {apple $\to$ orange};
        \node at (6.55, -0.65) {orange $\to$ apple};

        \node at (7.1,-2.6) {
            \includegraphics[width=0.13\textwidth,trim={.3cm 0 0 0}, clip]{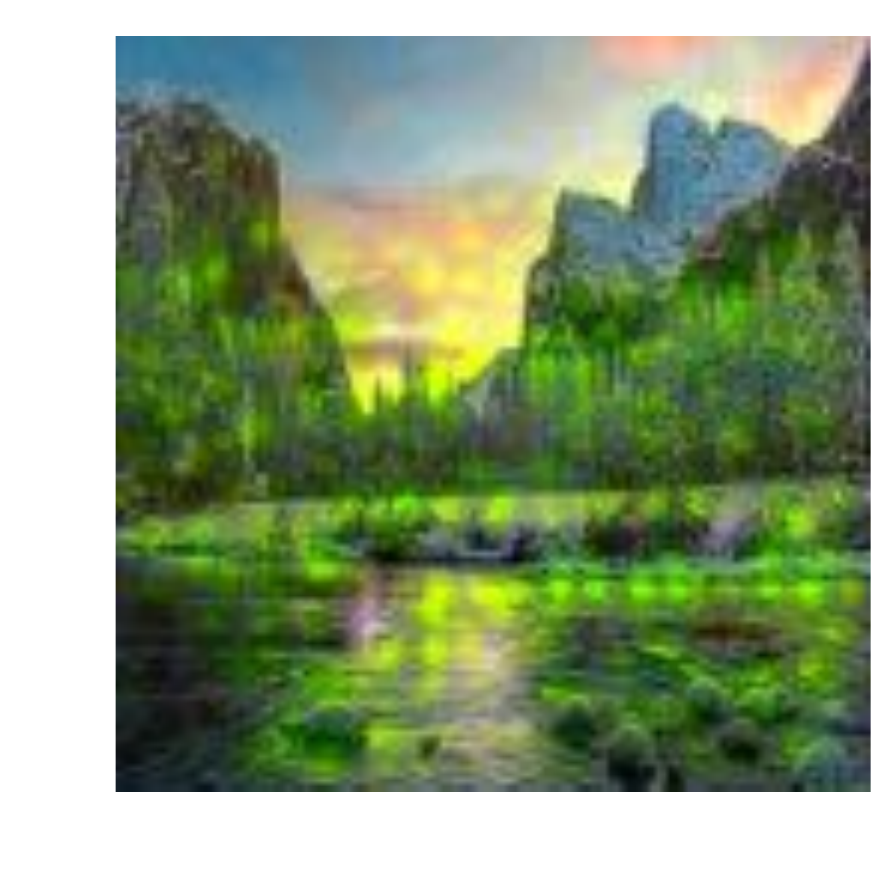}};
        \node at (6.1,-2.6) {
            \includegraphics[width=0.13\textwidth,trim={.3cm 0 0 0}, clip]{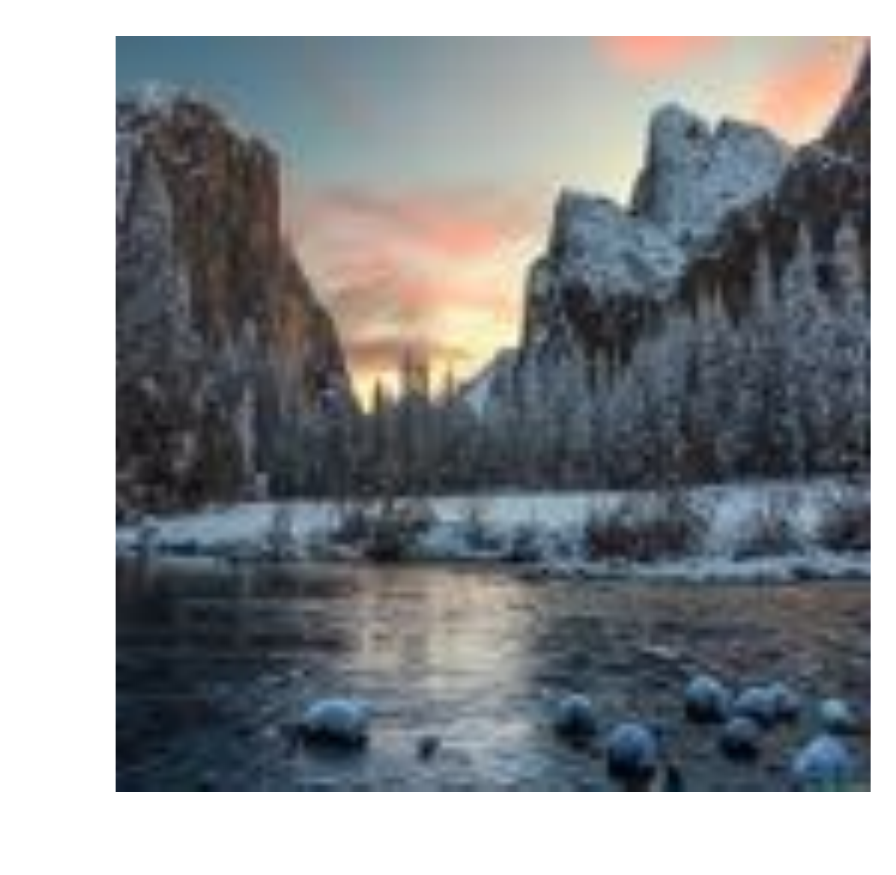}};
        \node at (5.1,-2.6) {
            \includegraphics[width=0.13\textwidth,trim={.3cm 0 0 0}, clip]{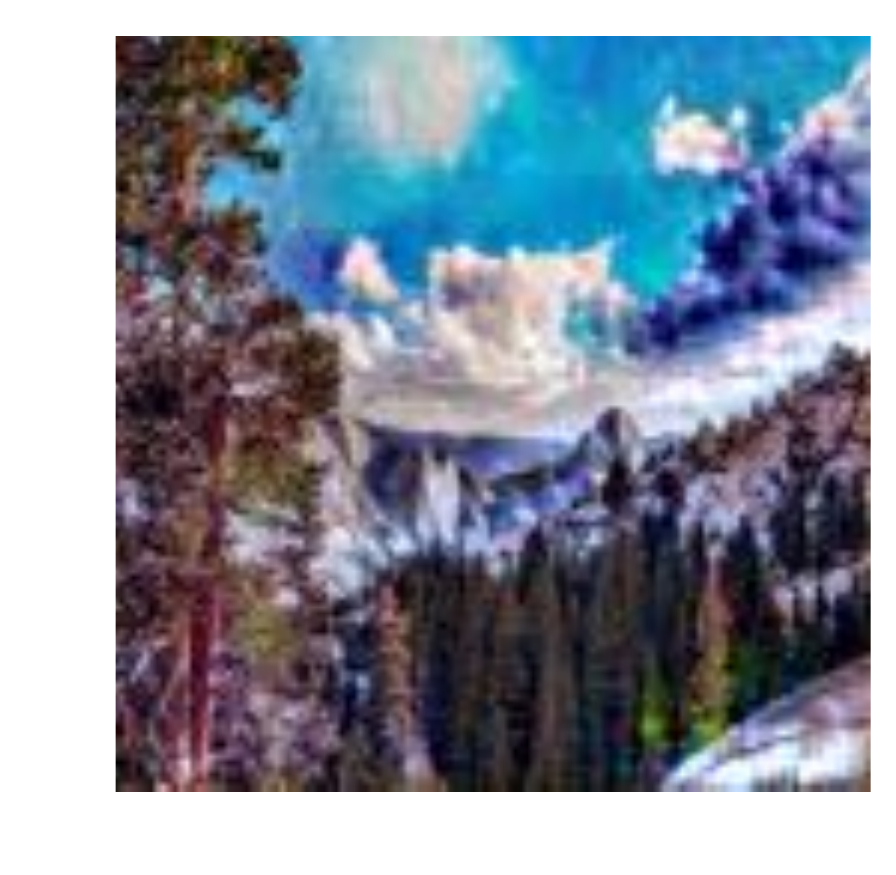}};
        \node at (4.1,-2.6) {
            \includegraphics[width=0.13\textwidth,trim={.3cm 0 0 0}, clip]{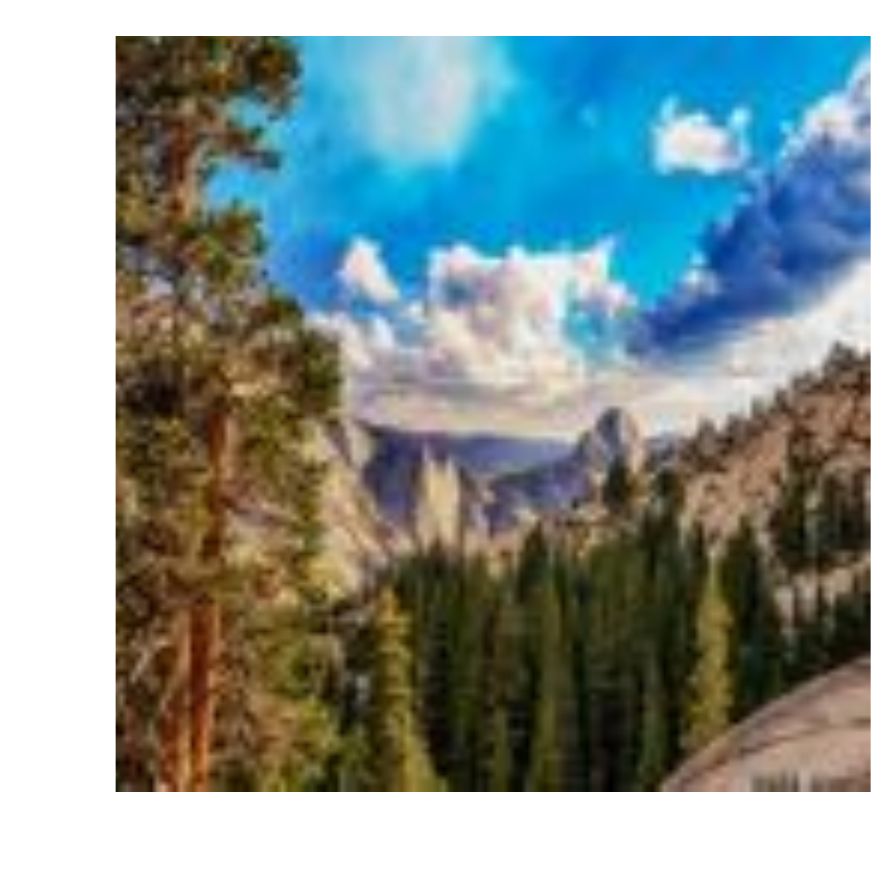}};
        \node at (3,-2.6) {
            \includegraphics[width=0.13\textwidth,trim={.3cm 0 0 0}, clip]{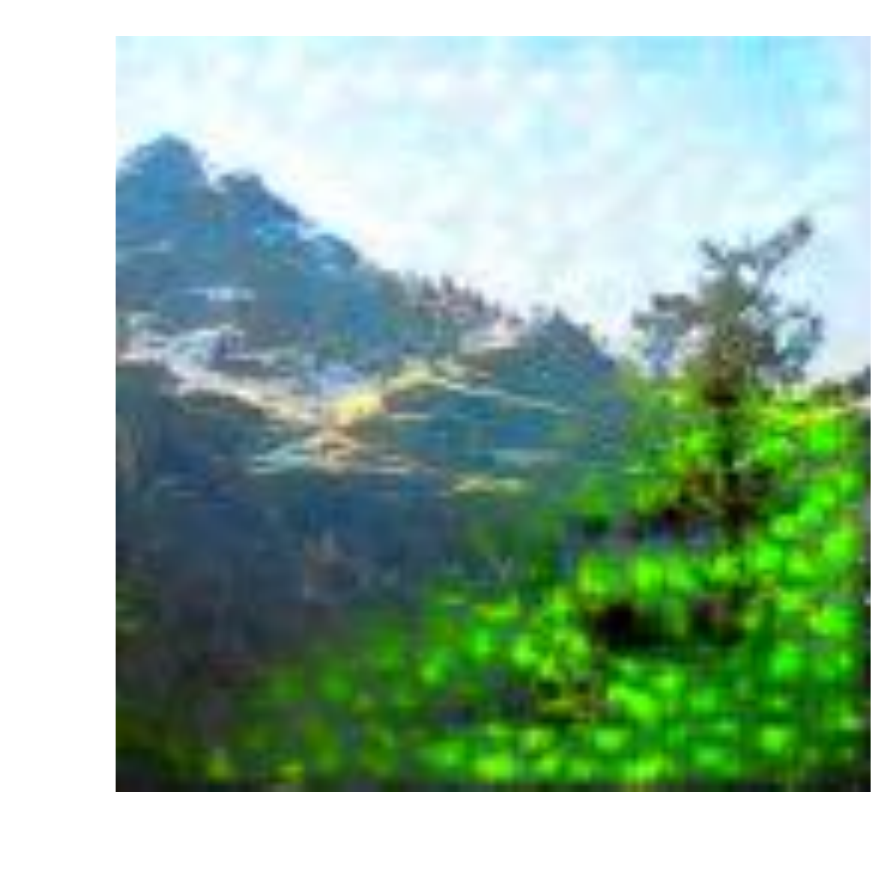}};
        \node at (2,-2.6) {
            \includegraphics[width=0.13\textwidth,trim={.3cm 0 0 0}, clip]{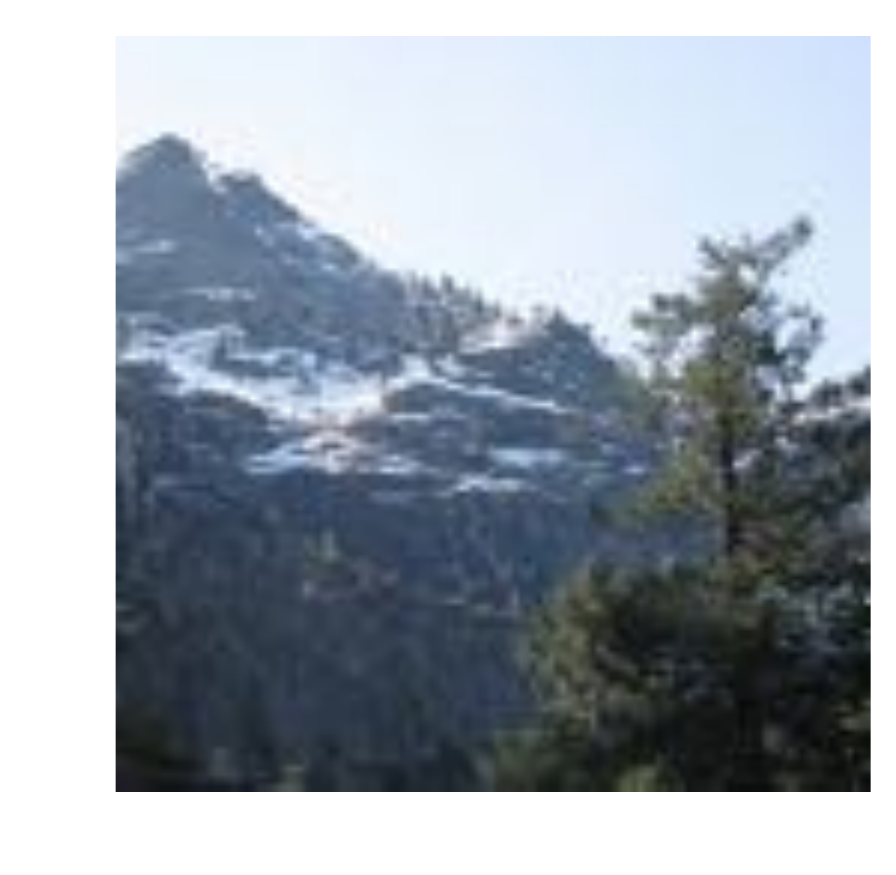}};
        \node at (1,-2.6) {
            \includegraphics[width=0.13\textwidth,trim={.3cm 0 0 0}, clip]{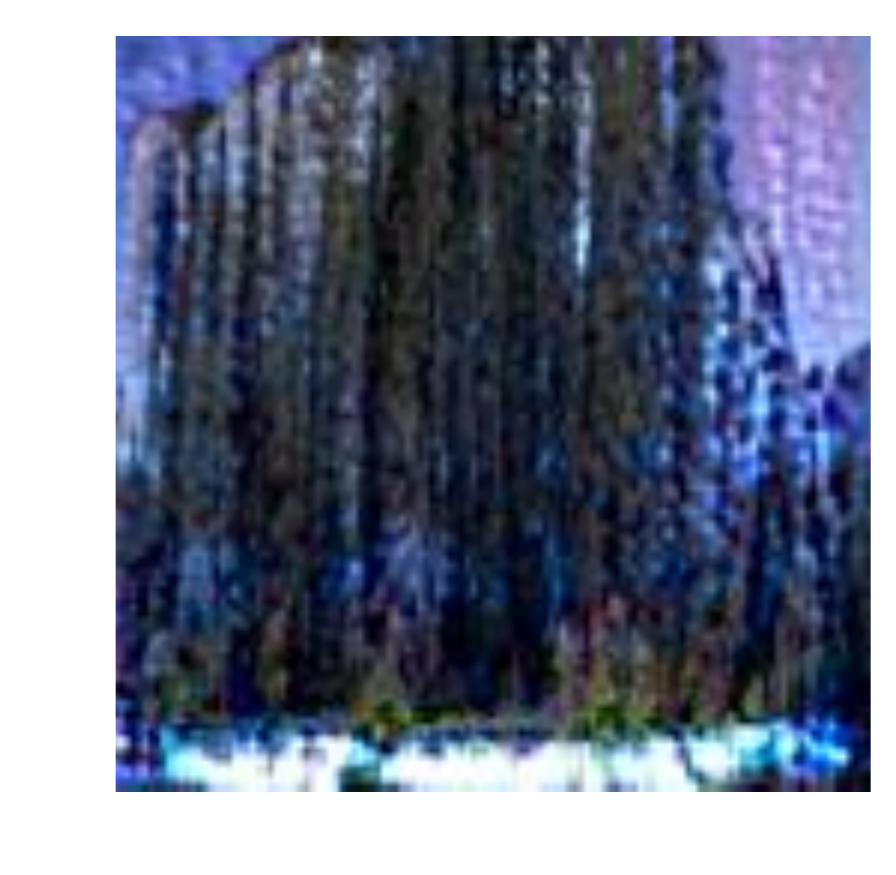}};
        \node at (0,-2.6) {
            \includegraphics[width=0.13\textwidth,trim={.3cm 0 0 0}, clip]{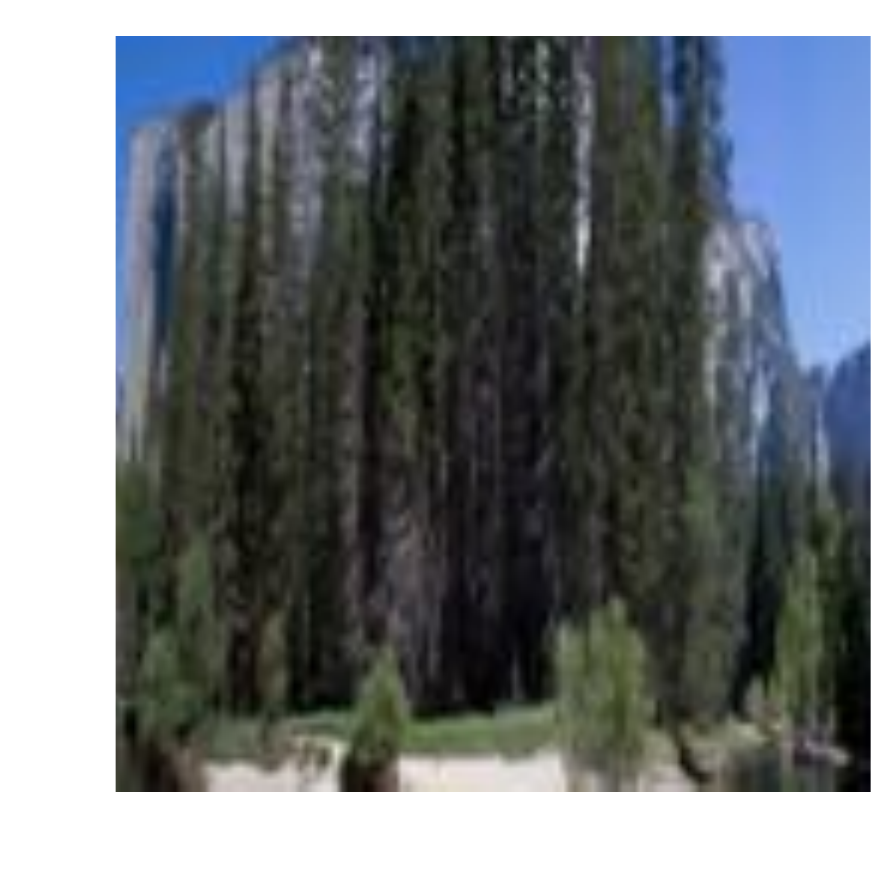}};

        \node at (0.5, -1.95) {summer $\to$ winter};
        \node at (2.5, -1.95) {winter $\to$ summer};
        \node at (4.55, -1.95) {summer $\to$ winter};
        \node at (6.55, -1.95) {winter $\to$ summer};

        \draw[dashed, very thick] (3.59, .8) -- (3.59, -3.15);
        \node at (1.5, -3.3) {(a) \emph{random} samples};
        \node at (5.5, -3.3) {(b) select samples};

    \end{tikzpicture}
    \end{center}
    \caption{Image-to-image translation on the {\HtoZ}, {\AtoO},
    and {\StoW} datasets~\cite{zhu2017unpaired} using PGD on the input of an
    $\ell_2$-robust model trained on that dataset.
    See Appendix~\ref{app:setup} for experimental details and 
    Figure~\ref{fig:h2z_app} for additional input-output pairs.}
    \label{fig:h2z}
\end{figure}
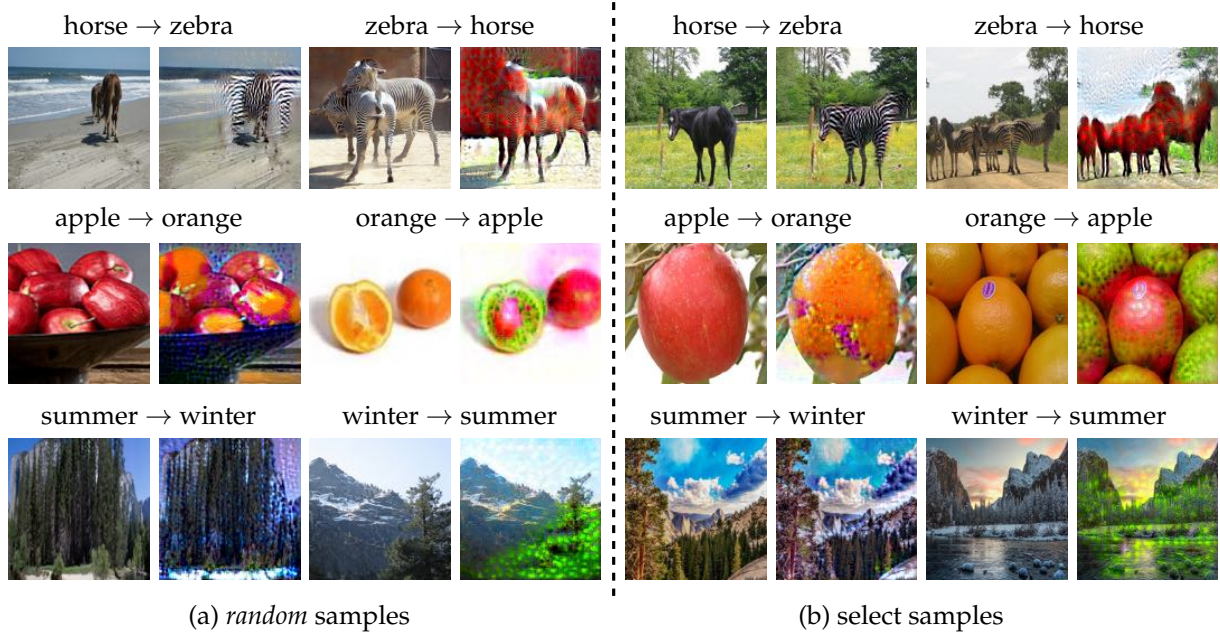

In Figure~\ref{fig:h2z}, we provide sample translations produced by our 
approach using robust models---each trained only on the source and 
target domains for the {\HtoZ}, {\AtoO}, and {\StoW} 
datasets~\cite{zhu2017unpaired} respectively. (For completeness,
we present in Appendix~\ref{app:omitted} Figure~\ref{fig:h2z_IN} 
results corresponding to using
a classifier trained on the complete ImageNet dataset.)
In general, we find that this procedure yields meaningful
translations by directly modifying characteristics of the image 
that are strongly tied to the corresponding domain (e.g., color, 
texture, stripes).

Note that, in order to manipulate such features, the model must have learned
them in the first place---for example, we want models to distinguish between
horses and zebras based on salient features such as stripes. For overly simple
 tasks, models might extract little
salient information (e.g., by relying on backgrounds instead of objects\footnote{In
    fact, we encountered such an issue with $\ell_\infty$-robust classifiers for
horses and zebras (Figure \ref{fig:fail_sky}). Note that generative
approaches also face similar issues, where the background is transformed instead
of the objects~\citep{zhu2017unpaired}.})
in which case our approach would not lead to meaningful translations.
Nevertheless, this not a fundamental barrier and
can be addressed by training on richer, more challenging datasets.
From this perspective, scaling to larger datasets (which can be difficult for
state-of-the-art methods such as GANs) is actually easy and advantageous for our approach.

\paragraph{Unpaired datasets.} Datasets for translation tasks
often comprise source-target domain pairs~\cite{isola2017image}.
For such datasets, the task can be straightforwardly cast into a supervised
learning framework.
In contrast, our method operates in the {\em unpaired} setting, where
samples from the source and target domain are provided without an
explicit pairing~\cite{zhu2017unpaired}.
This is due to the fact that our method only requires a classifier capable
of distinguishing between the source and target domains.

%% file: superresolution.tex
Super-resolution refers to the task of recovering high-resolution images given 
their low resolution version~\cite{dabov2007video,burger2012image}.
While this goal is underspecified, our aim is
to produce a high-resolution image that is consistent with the input and
plausible to a human. 

In order to adapt our framework to this problem,
we cast super-resolution as the task of accentuating the salient
features of low-resolution images.
This can be achieved by maximizing the score predicted by a robust classifier 
(trained on the original high-resolution dataset)
for the underlying class.
At the same time, to ensure that the structure and high-level content is preserved, we penalize
large deviations from the original low-resolution image.
Formally, given a robust classifier
and a low-resolution image $x_L$ belonging to class
$y$, we  use PGD to solve
\begin{equation}
    \hat{x}_H = \argmin_{||x' - \uparrow(x_{L})|| < \eps} \; \mathcal{L}(x', y)
\label{eq:superres}
\end{equation}
where $\uparrow(\cdot)$ denotes the up-sampling operation based on nearest
neighbors, and $\eps$ is a small constant.

\begin{figure}[!h]
	\centering
	\begin{subfigure}[b]{0.495\textwidth}
		\centering
		\includegraphics[width=1\textwidth]{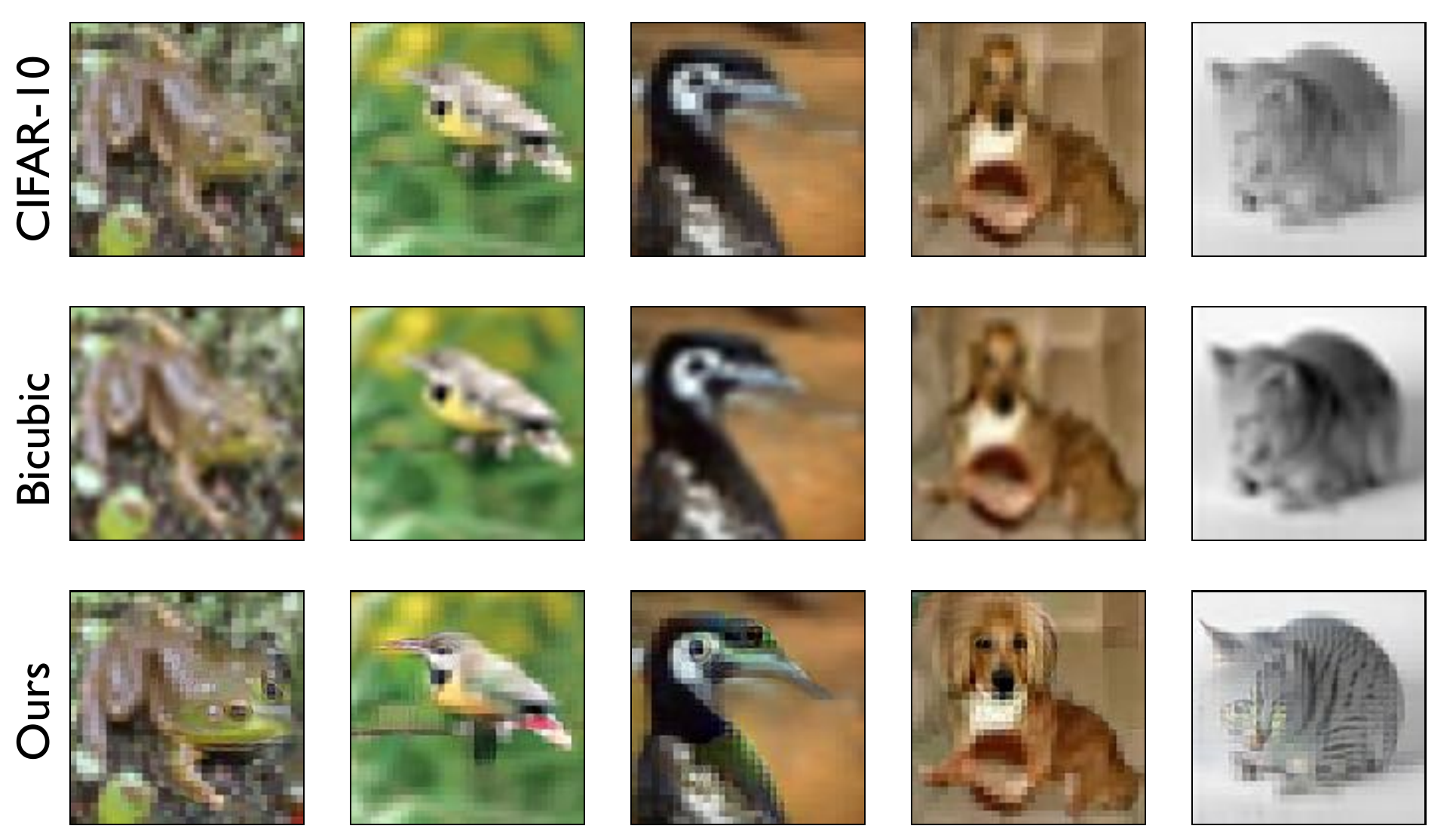}
		\caption{$7$x super-resolution on CIFAR-10}
		\label{fig:superres_cifar}
	\end{subfigure}
	\hfil
	\begin{subfigure}[b]{0.495\textwidth}
		\centering
		\includegraphics[width=1\textwidth]{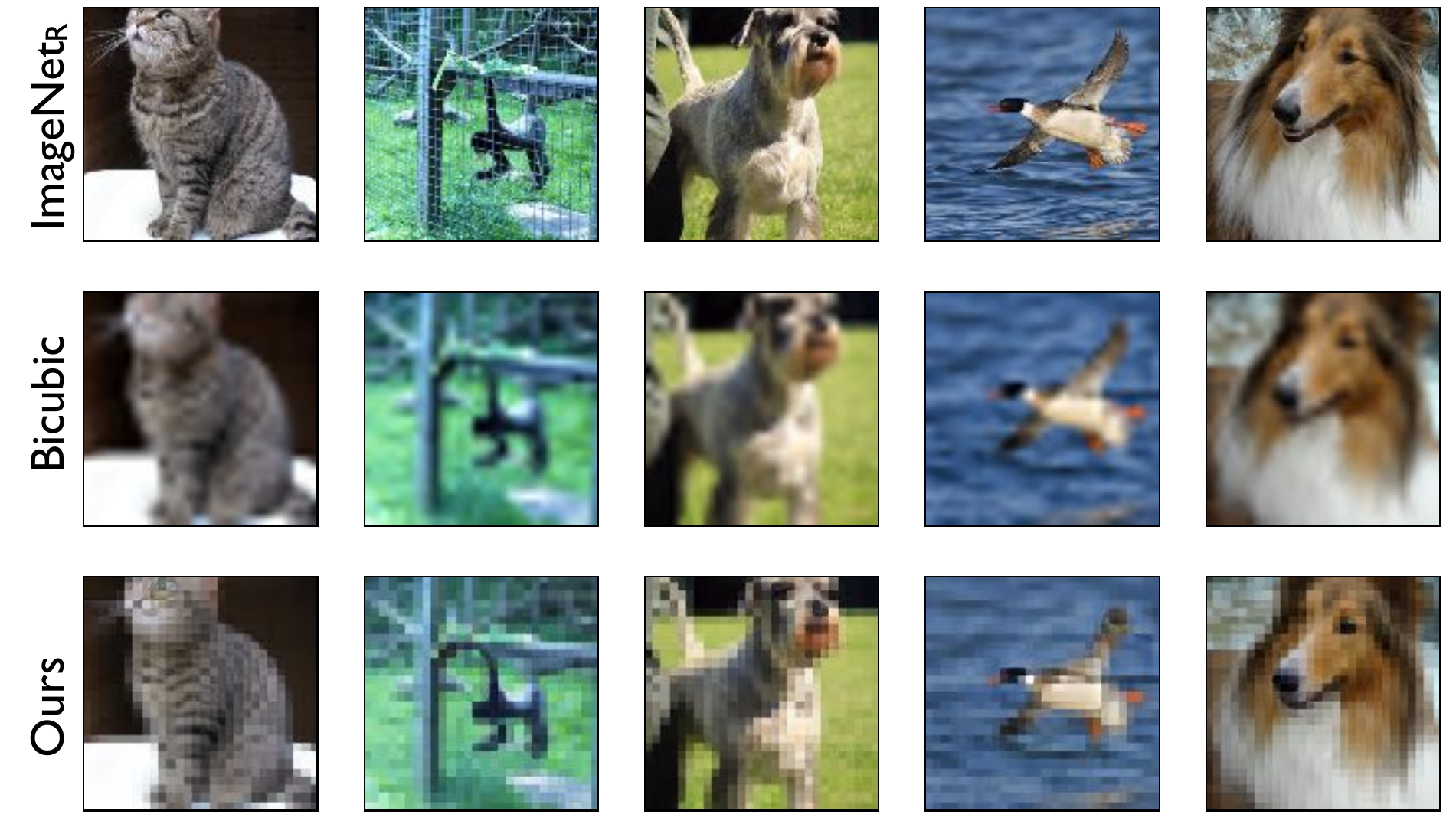}
		\caption{$8$x super-resolution on restricted ImageNet}
		\label{fig:superres_in}
	\end{subfigure}
	\caption{Comparing approaches for super-resolution. 
		\textit{Top:} \emph{random} samples from the test set;
		\textit{middle:} upsampling using bicubic interpolation; and
		\textit{bottom:} super-resolution using robust models. We obtain 
		semantically meaningful reconstructions that are especially sharp
		in regions that contain class-relevant information.
	}
	\label{fig:superres}
\end{figure}

We use this approach to upsample
\emph{random} $32\times32$ CIFAR-10 images to full ImageNet size ($224 
\times 224$)---cf. Figure~\ref{fig:superres_cifar}. For comparison, we also show upsampled images obtained 
from bicubic interpolation. 
In Figure~\ref{fig:superres_in}, we visualize the results for super-resolution on 
\emph{random} $8$-fold down-sampled images from the restricted 
ImageNet dataset. Since  in the latter case we have access to ground truth 
high-resolution images (actual dataset samples), we can
compute the Peak Signal-to-Noise Ratio (PSNR) of the reconstructions.
Over the Restricted ImageNet 
test set, our approach yields a PSNR of $21.53$ 
($95$\% CI [$21.49$, $21.58$]) compared to
$21.30$ ($95$\% CI [$21.25$, $21.35$]) from bicubic interpolation. 
In general, our approach produces 
high-resolution samples that are substantially sharper, particularly in regions 
of the image that contain salient class information.

Note that the pixelation of the resulting images can be attributed to using a
very crude upsampling of the original, low-resolution image as a starting point for our
optimization. Combining this method with a more sophisticated initialization
scheme (e.g., bicubic interpolation) is likely to yield better overall
results.

%% file: robpaint.tex
Recent work has explored building deep learning--based interactive tools for
image synthesis and manipulation.
For example, GANs have been used to transform simple
sketches~\cite{chen2018sketchygan,park2019semantic} into realistic images.
In fact, recent work has pushed this one step further by building a tool
that allows object-level composition of scenes using GANs~\cite{bau2019gan}.
In this section, we show how our framework can be used to enable similar artistic applications.

\begin{figure}[!h]
\includegraphics[width=\textwidth]{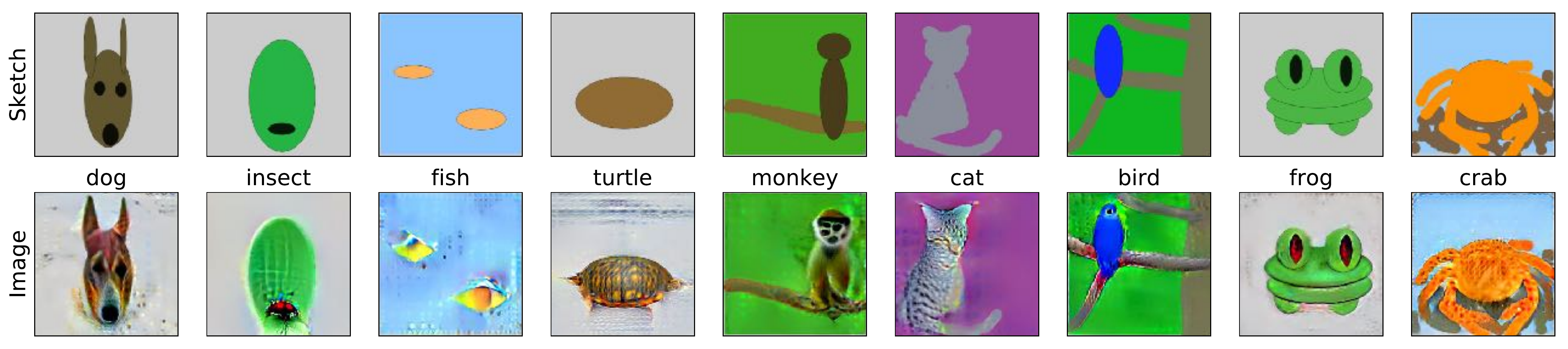}
\caption[]{Sketch-to-image using robust model gradients. \emph{Top:} 
manually drawn sketches of animals; and \emph{bottom:} result of 
performing PGD towards a chosen class. The resulting images appear
realistic looking while preserving key characteristics of the original 
sketches\footnotemark.}
\label{fig:aipaint}
\end{figure}

\paragraph{Sketch-to-image.} By performing PGD to maximize the probability of 
a chosen target class, we can use robust models to convert 
hand-drawn sketches to natural images.  The resulting images (Figure~\ref{fig:aipaint}) 
appear realistic and contain fine-grained characteristics of the corresponding 
class.

\footnotetext{Sketches were produced by a graduate student without any
training in arts.}

\paragraph{Feature Painting.}
Generative model--based paint applications often allow the user to control more
fine-grained features, as opposed to just the overall class. We now show that we
can perform similar feature manipulation through a minor modification to our basic primitive
of class score maximization. Our methodology is based on an observation of 
\citet{engstrom2019learning},
wherein manipulating individual activations within representations\footnote{We refer 
	to the pre-final layer of a network as the representation layer. Then, the 
	network prediction can simply be viewed as the
	output of a linear classifier on the representation.} 
of a robust model actually results in consistent and meaningful 
changes to high-level image features (e.g., adding stripes to objects).  
We can thus build a tool to paint specific features onto images by
maximizing individual activations directly, instead of just the class scores.

Concretely, given an
image $x$, if we want to add a single feature corresponding to component $f$ of
the representation vector $R (x)$ in the region corresponding to a binary mask
$m$, we simply apply PGD to solve
\begin{equation}
x_I = {\argmax}_{x'}\ R(x')_f - \lambda_P || (x - x') \odot (1 - m)||.
\label{eq:ganpaint}
\end{equation}
\begin{figure}[!h]
	\includegraphics[width=\textwidth]{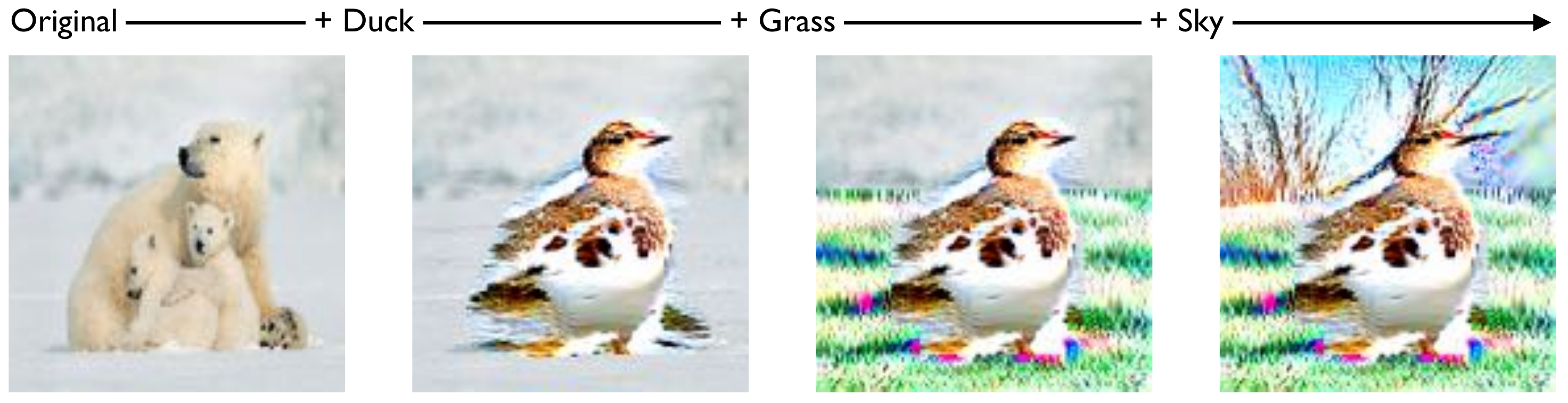}
	\caption{Paint-with-features using a robust model---we present a
	sequence of images obtained by successively adding specific features to 
    select regions of the image by solving \eqref{eq:ganpaint}.}
	\label{fig:robpaint}
\end{figure}
In Figure~\ref{fig:robpaint}, we demonstrate progressive addition of features at various levels of
granularity (e.g., grass or sky) to selected regions of the input image. We can observe that
such direct maximization of individual activations gives rise to a versatile
paint tool.

%% file: conclusion.tex
In this work, we leverage the basic classification framework to perform a wide
range of image synthesis tasks. In particular, we find that the features learned
by a basic classifier are sufficient for {\em all} these tasks, provided this
classifier is {\em adversarially robust}. We then show how this insight gives
rise to a versatile toolkit that is simple, reliable, and
straightforward to extend to other large-scale datasets. This is in stark contrast
to state-of-the-art
approaches~\cite{goodfellow2014generative,karras2018progressive,brock2019large}
which typically rely on architectural, algorithmic, and task-specific
optimizations to succeed at
scale~\cite{salimans2016improved,daskalakis2018training,miyato2018spectral}. In
fact, unlike these approaches, our methods actually {\em benefit} from scaling
to more complex datasets---whenever the underlying classification task is rich
and challenging, the classifier is likely to learn more fine-grained features.

We also note that throughout this work, we choose to employ the most minimal
version of our toolkit. In particular, we refrain from using extensive tuning or
task-specific optimizations. This is intended to demonstrate the potential of
our core framework itself, rather than to exactly match/outperform the state of
the art. We fully expect that better training methods, improved notions of
robustness, and domain knowledge will yield even better results.

More broadly, our findings suggest that adversarial robustness might be a
property that is desirable beyond security and reliability contexts. Robustness
may, in fact, offer a path towards building a more human-aligned machine
learning toolkit.

%% file: acks.tex
We thank Chris Olah for helpful pointers to related work in class visualization.

Work supported in part by the NSF grants CCF-1553428, CCF-1563880,
CNS-1413920, CNS-1815221, IIS-1447786, IIS-1607189, the Microsoft Corporation, the Intel Corporation,
the MIT-IBM Watson AI Lab research grant, and an Analog Devices Fellowship.

%% file: setup.tex
\subsection{Datasets}
\label{sec:dss}
For our experimental analysis, we use the CIFAR-10~\cite{krizhevsky2009learning} 
and ImageNet~\cite{russakovsky2015imagenet} datasets. Since obtaining a robust
classifier for the full ImageNet dataset is known to be a challenging and
computationally expensive problem, we also conduct experiments 
on a  ``restricted'' version if the ImageNet dataset with 9 super-classes 
shown in Table~\ref{tab:classes}. For image translation we use the
{\HtoZ}, {\AtoO}, and {\StoW} datasets~\cite{zhu2017unpaired}.

\begin{table}[!h]
	\caption{Classes used in the Restricted ImageNet model. The class ranges are
		inclusive.}
	\begin{center}
		\begin{tabular}{ccc}
			\toprule
			\textbf{Class} & \phantom{x} & \textbf{Corresponding ImageNet Classes} \\
			\midrule
			``Dog'' &&   151  to 268    \\ 
			``Cat'' &&   281  to 285    \\
			``Frog'' &&   30  to 32    \\
			``Turtle'' &&   33  to 37    \\
			``Bird'' &&   80  to 100    \\
			``Primate'' &&   365  to 382    \\
			``Fish'' &&   389  to 397    \\
			``Crab'' &&   118  to 121    \\
			``Insect'' &&   300  to 319    \\
			\bottomrule
		\end{tabular}
	\end{center}
	\label{tab:classes}
\end{table}

\subsection{Models}
\label{sec:models}
We use the standard ResNet-50 architecture~\cite{he2016deep} for our 
adversarially trained classifiers on all datasets. Every model is trained with 
data augmentation, momentum of $0.9$ and weight decay of $5e^{-4}$.
Other hyperparameters are provided in Tables~\ref{tab:nat_hyper} 
and~\ref{tab:adv_hyper}. 

\begin{table}[!h]
	\caption{Standard hyperparameters for the models trained in the main paper.}
	\begin{center}
		\begin{tabular}{l|cccc}
			\toprule
			{\bf Dataset} & Epochs & LR & Batch Size & LR Schedule \\
			\midrule
			CIFAR-10 & 350 & 0.01 & 256 & Drop by 10 at epochs $\in [150, 250]$  \\
			restricted ImageNet & 110 & 0.1 & 128 & Drop by 10 at epochs $\in [30, 60]$  \\
			ImageNet & 110 &  0.1 & 256 &  Drop by 10 at epochs $\in [100]$   \\
			\HtoZ & 350 & 0.01 & 64 & Drop by 10 at epochs $\in [50, 100]$  \\
			\AtoO & 350 & 0.01 & 64 & Drop by 10 at epochs $\in [50, 100]$  \\
			\StoW & 350 & 0.01 & 64 & Drop by 10 at epochs $\in [50, 100]$  \\
			\bottomrule
		\end{tabular}
	\end{center}
	\label{tab:nat_hyper}
\end{table}

\subsection{Adversarial training}
In all our experiments, we train robust classifiers by employing the adversarial 
training methodology~\cite{madry2018towards} with an $\ell_2$ perturbation 
set. The hyperparameters used for robust training of each of our models
are provided in Table~\ref{tab:adv_hyper}.

\begin{table}[!h]
	\caption{Hyperparameters used for adversarial training.}
\begin{center}
\setlength{\tabcolsep}{.8cm}
\begin{tabular}{cccc}
\toprule 
Dataset & $\epsilon$ & \# steps & Step size \\
\midrule
CIFAR-10 & 0.5 & 7 & 0.1 \\
restricted ImageNet & 3.5 & 7 & 0.1 \\
ImageNet & 3 & 7 & 0.5 \\
{\HtoZ} & 5 & 7 & 0.9 \\
{\AtoO} & 5 & 7 & 0.9 \\
{\StoW} & 5 & 7 & 0.9 \\
\bottomrule 
\end{tabular}
\end{center}
	\label{tab:adv_hyper}
\end{table}

\subsection{Note on hyperparameter tuning}
\label{app:tune}
Note that we did not perform \emph{any} hyperparameter tuning 
for the hyperparameters in Table~\ref{tab:nat_hyper} because of 
computational constraints. We use the relatively standard benchmark 
$\epsilon$ of 0.5 for CIFAR-10---the rest of the values of $\epsilon$ were chosen 
roughly by scaling this up by the appropriate constant (i.e. proportional
to sqrt(d))---we note that the networks are not critically sensitive to 
these values of epsilon (e.g. a CIFAR-10 model trained with $\epsilon=1.0$ 
gives almost the exact same results). Due to restrictions on compute we 
did not grid search over $\epsilon$, but finding a more direct manner 
in which to set $\epsilon$ (e.g. via a desired adversarial accuracy) is an 
interesting future direction.

\subsection{Targeted Attacks in Figure~\ref{fig:targeted}}

\begin{center}
\setlength{\tabcolsep}{.8cm}
\begin{tabular}{cccc}
\toprule 
Dataset & $\epsilon$ & \# steps & Step size \\
\midrule
restricted ImageNet & 300 & 500 & 1 \\
\bottomrule 
\end{tabular}
\end{center}

\subsection{Image-to-image translation}

\begin{center}
\setlength{\tabcolsep}{.8cm}
\begin{tabular}{cccc}
\toprule 
Dataset & $\epsilon$ & \# steps & Step size \\
\midrule
ImageNet & 60 & 80 & 1 \\
{\HtoZ} & 60 & 80 & 0.5 \\
{\AtoO} & 60 & 80 & 0.5 \\
{\StoW} & 60 & 80 & 0.5 \\
\bottomrule 
\end{tabular}
\end{center}

\subsection{Generation}
In order to compute the class conditional Gaussians for high resolution images
(224$\times$224$\times$3) we downsample the images by a factor of 4 and upsample
the resulting seed images with nearest neighbor interpolation.

\begin{center}
\setlength{\tabcolsep}{.8cm}
\begin{tabular}{cccc}
\toprule 
Dataset & $\epsilon$ & \# steps & Step size \\
\midrule
CIFAR-10 & 30 & 60 & 0.5 \\
restricted ImageNet & 40 & 60 & 1 \\
ImageNet & 40 & 60 & 1 \\
\bottomrule 
\end{tabular}
\end{center}

\subsubsection{Inception Score}
\label{app:is}
Inception score is computed based on 50k class-balanced samples from
each dataset using code provided in \url{https://github.com/ajbrock/BigGAN-PyTorch}.

\subsection{Inpainting}
To create a corrupted image, we select a patch of a given size at a random location in the image. We 
reset all pixel values in the patch to be the average pixel value over the entire image (per channel).
\begin{center}
	\setlength{\tabcolsep}{.8cm}
	\begin{tabular}{ccccc}
		\toprule 
		Dataset & patch size & $\epsilon$ & \# steps & Step size \\
		\midrule
		restricted ImageNet & 60 & 21 & 0.1 & 720 \\
		\bottomrule 
	\end{tabular}
\end{center}

\subsection{Super-resolution}

\begin{center}
	\setlength{\tabcolsep}{.8cm}
	\begin{tabular}{ccccc}
		\toprule 
		Dataset & $\uparrow$ factor & $\epsilon$ & \# steps & Step size \\
		\midrule
		CIFAR-10 & 7 & 15 & 1 & 50 \\
		restricted ImageNet & 8 & 8 & 1 & 40 \\
		\bottomrule 
	\end{tabular}
\end{center}

%% file: omitted.tex
\begin{figure}[!h]
\begin{center}
    {\HtoZ} \\[.2cm]
    \includegraphics[width=1\textwidth]{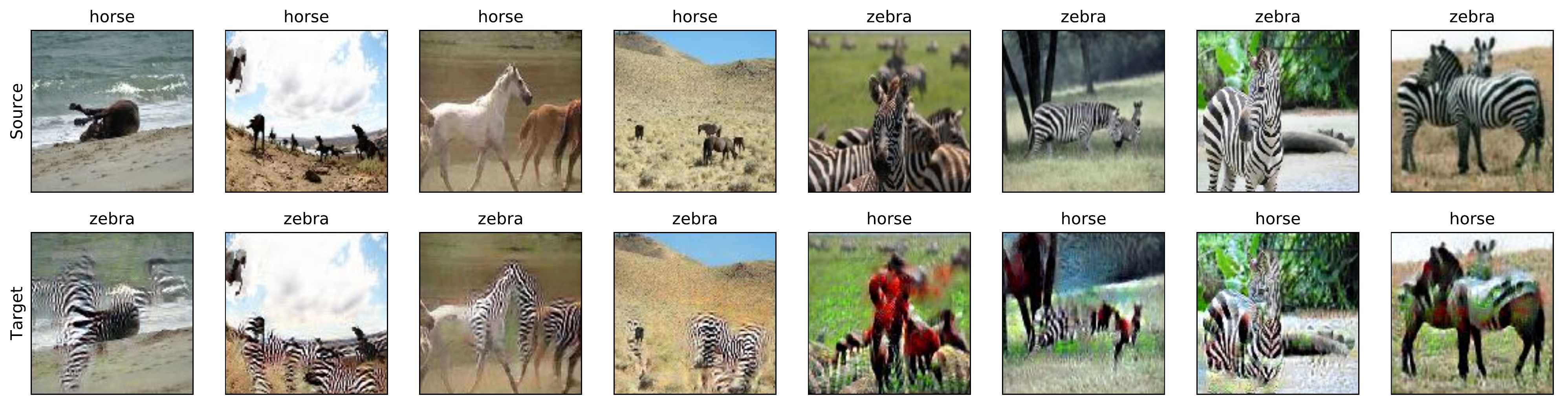} \\[.3cm]
    {\AtoO} \\[.2cm]
    \includegraphics[width=1\textwidth]{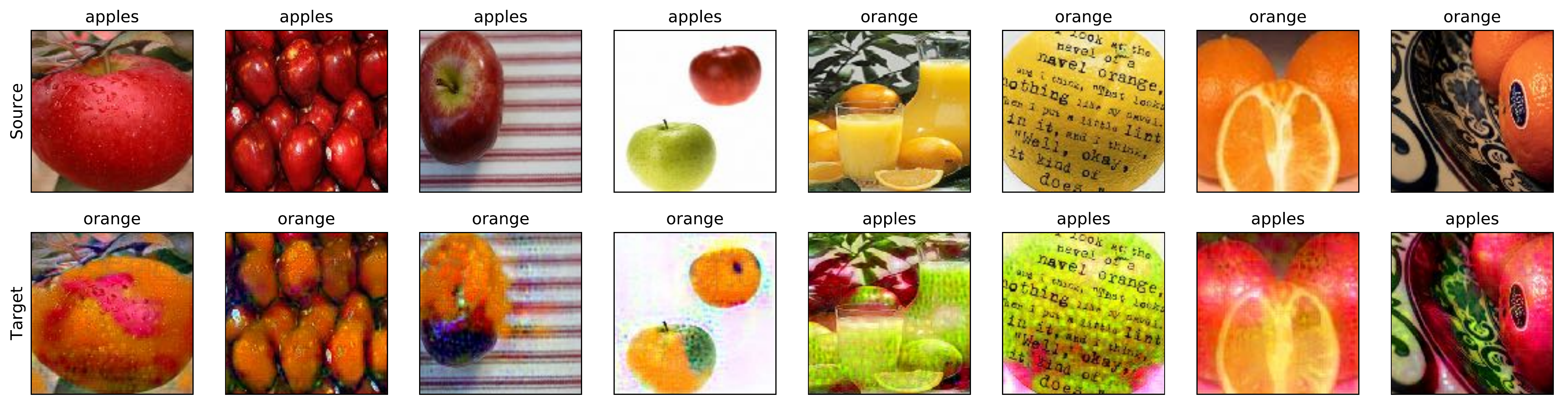} \\[.3cm]
    {\StoW} \\[.2cm]
    \includegraphics[width=1\textwidth]{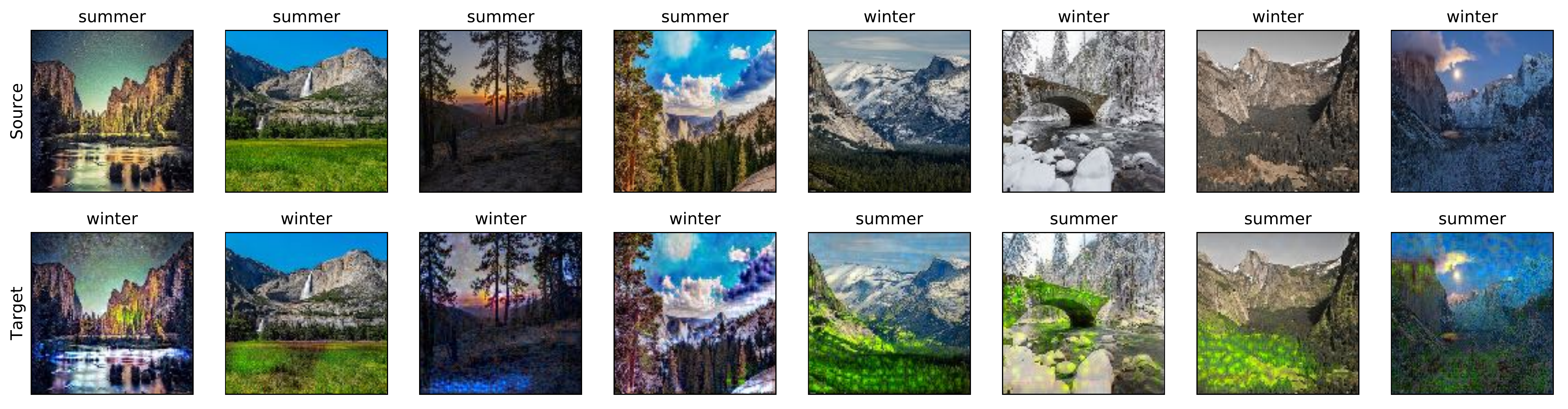}
\end{center}
\caption{Random samples for image-to-image translation on the {\HtoZ}, {\AtoO},
    and {\StoW} datasets~\cite{zhu2017unpaired}. Details in
    Appendix~\ref{app:setup}.}
\label{fig:h2z_app}
\end{figure}

\begin{figure}[!h]
\begin{center}
    \setlength{\tabcolsep}{0.1cm}
    \begin{tabular}{cc}
        Horse $\to$ Zebra & Apple $\to$ Orange \\
        \includegraphics[width=.48\textwidth]{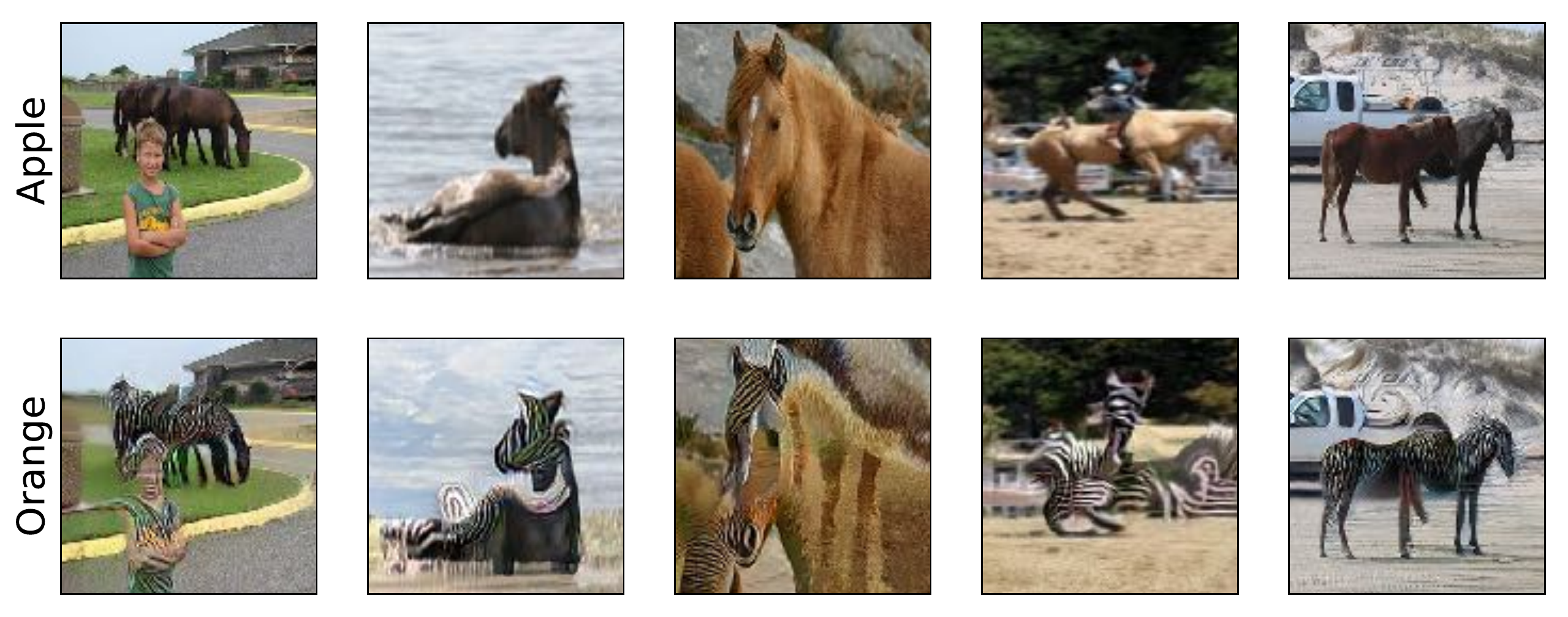} &
        \includegraphics[width=.48\textwidth]{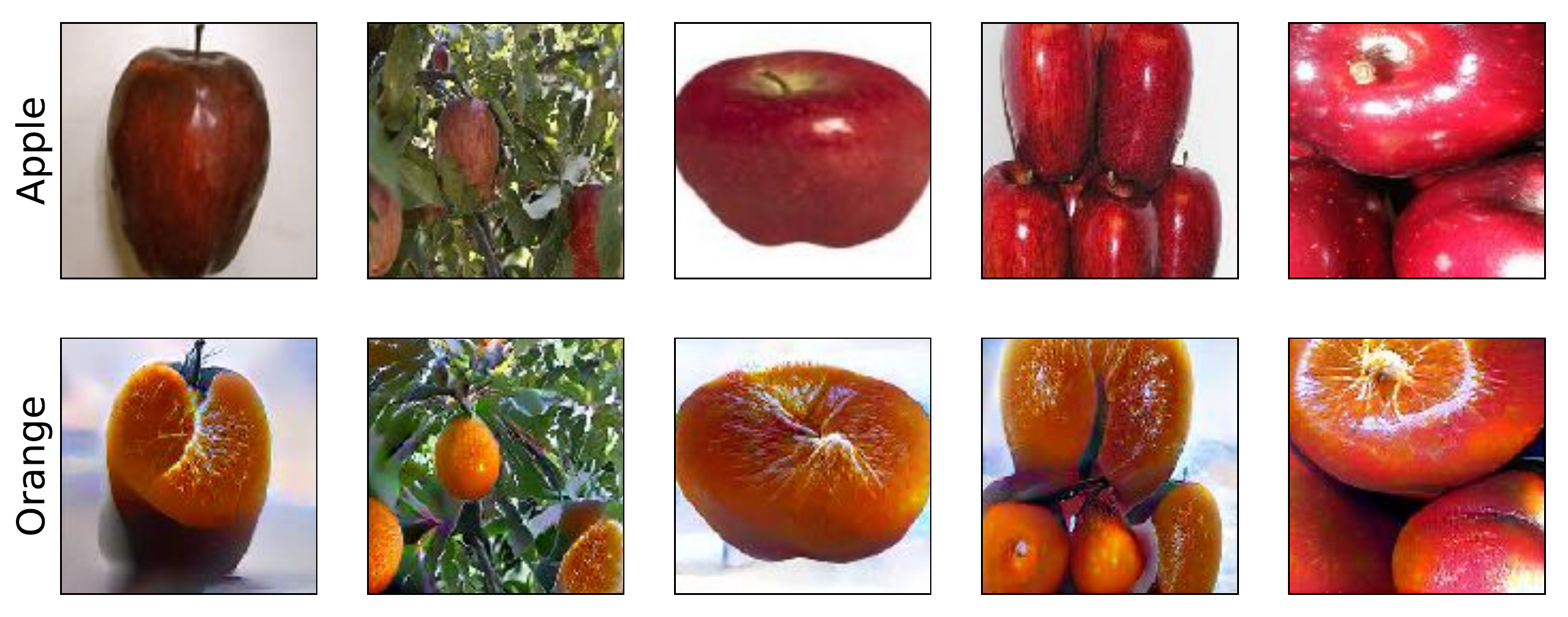} 
    \end{tabular}
\end{center}
\caption{Random samples for image-to-image translation on the {\HtoZ} and
    {\AtoO} datasets~\cite{zhu2017unpaired} using the {\em same} robust model
trained on the {\em entire ImageNet} dataset. Here we use ImageNet classes
``zebra'' (340) and ``orange'' (950).}
\label{fig:h2z_IN}
\end{figure}

\begin{figure}[!h]
\begin{center}
    \includegraphics[width=.95\textwidth]{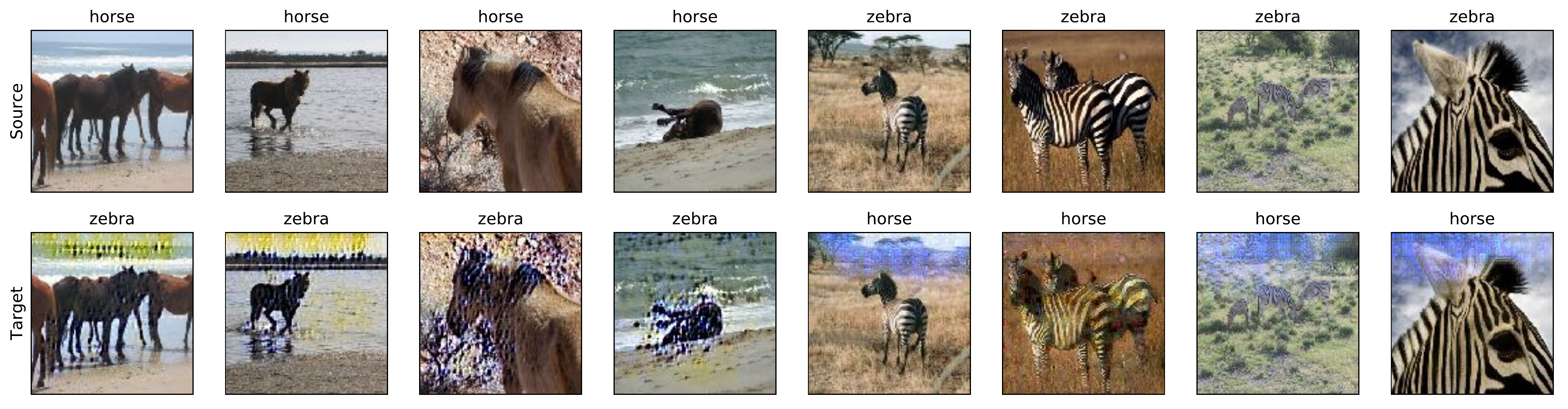}
\end{center}
\caption{Training an $\ell_\infty$-robust model on the {\HtoZ} dataset does not
lead to plausible image-to-image translation. The model appears to associate
``horse'' with ``blue sky'' in which case the zebra to horse translation does
not behave as expected.}
\label{fig:fail_sky}
\end{figure}

\begin{figure}[!ht]
	\centering
	\includegraphics[width=1\textwidth]{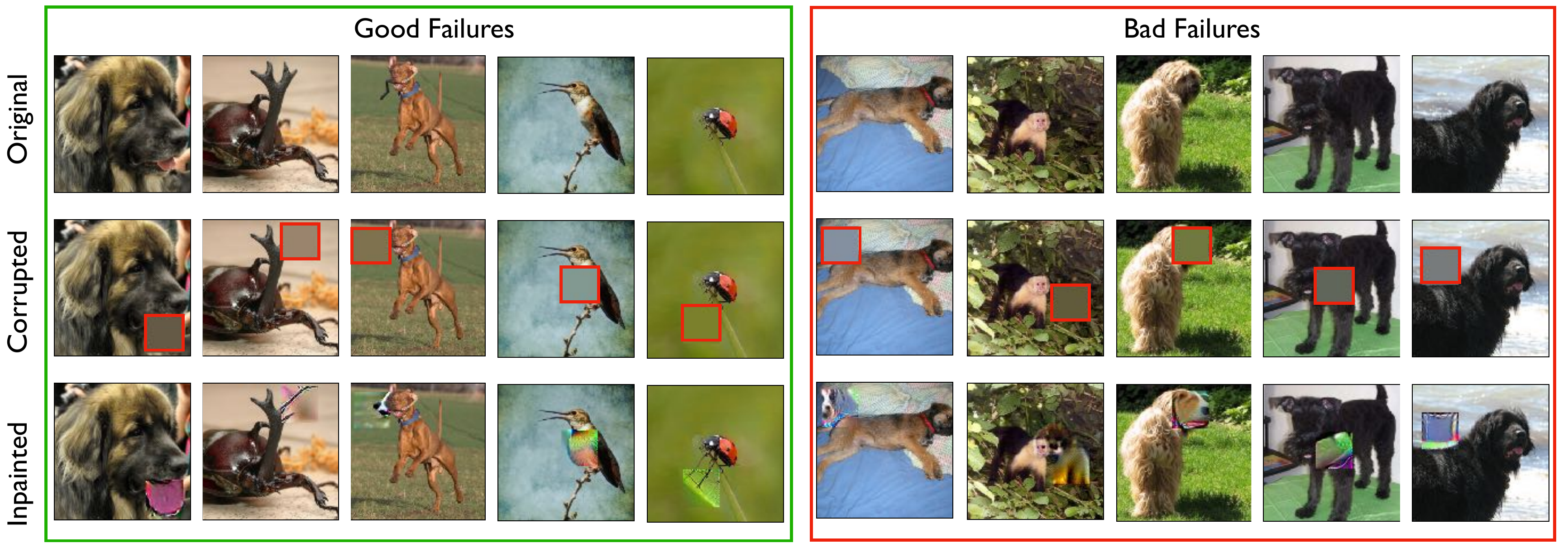}
	\caption{Failure cases for image inpainting using robust models -- 
		\textit{top:} original, \textit{middle:} corrupted and \textit{bottom:} 
		inpainted samples. To recover 
		missing regions, we use PGD to maximise the class score of 
		the image under a robust model while penalizing changes to 
		the uncorrupted regions. The failure modes can be categorized
	    into ``good'' failures -- where the infilled region is semantically
        consistent with the rest of the image but differs from the original;
       and ``bad'' failures --  where the inpainting is clearly erroneous to
       a human.}
	\label{fig:inpainting_bloopers}
\end{figure}

\begin{figure}[!htp]
\begin{center}
    \includegraphics[width=\textwidth]{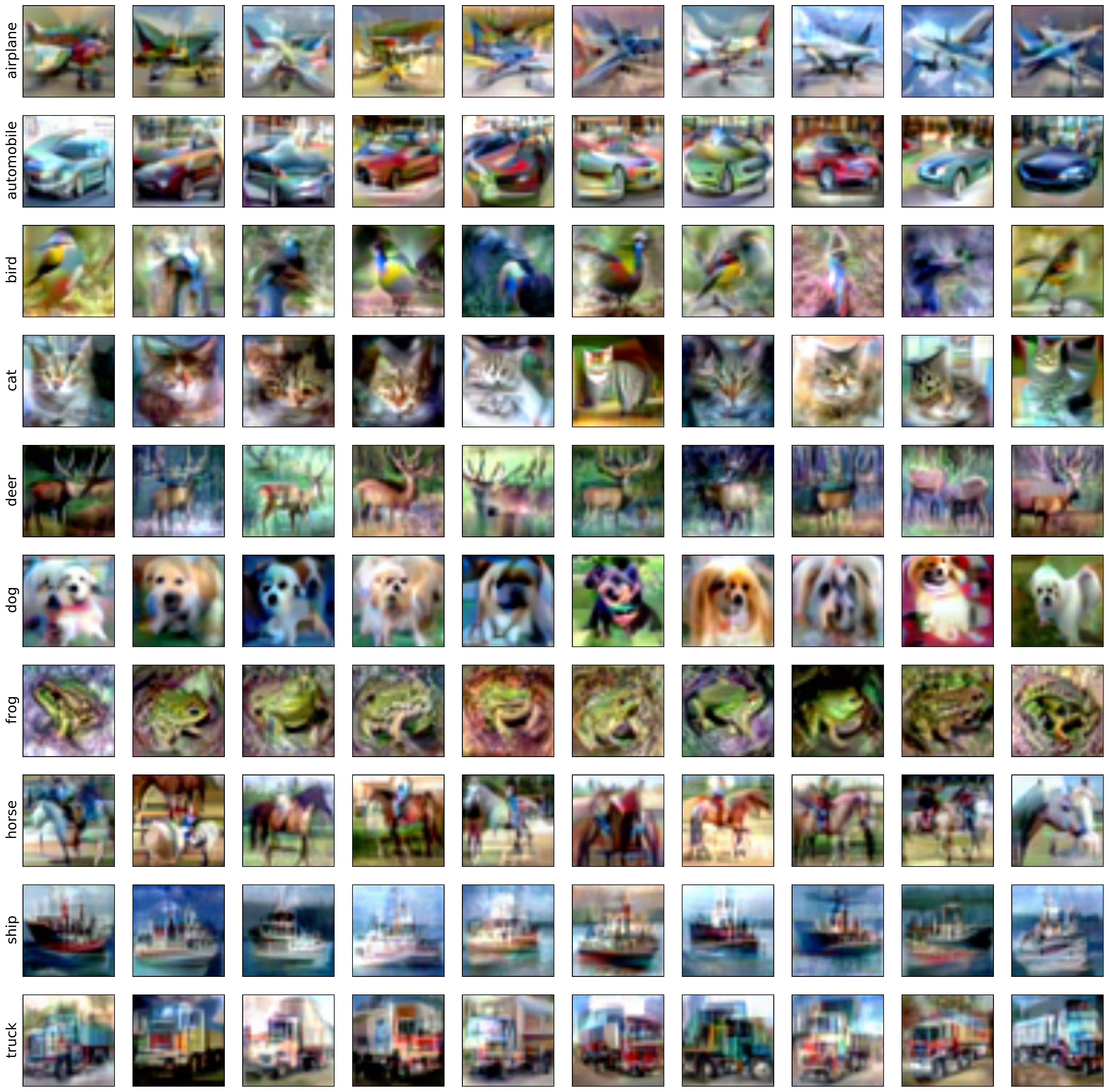}
\end{center}
\caption{Random samples generated for the CIFAR dataset.}
\label{fig:CIFAR_gen_full}
\end{figure}

\begin{figure}[!htp]
\begin{center}
    \includegraphics[width=\textwidth]{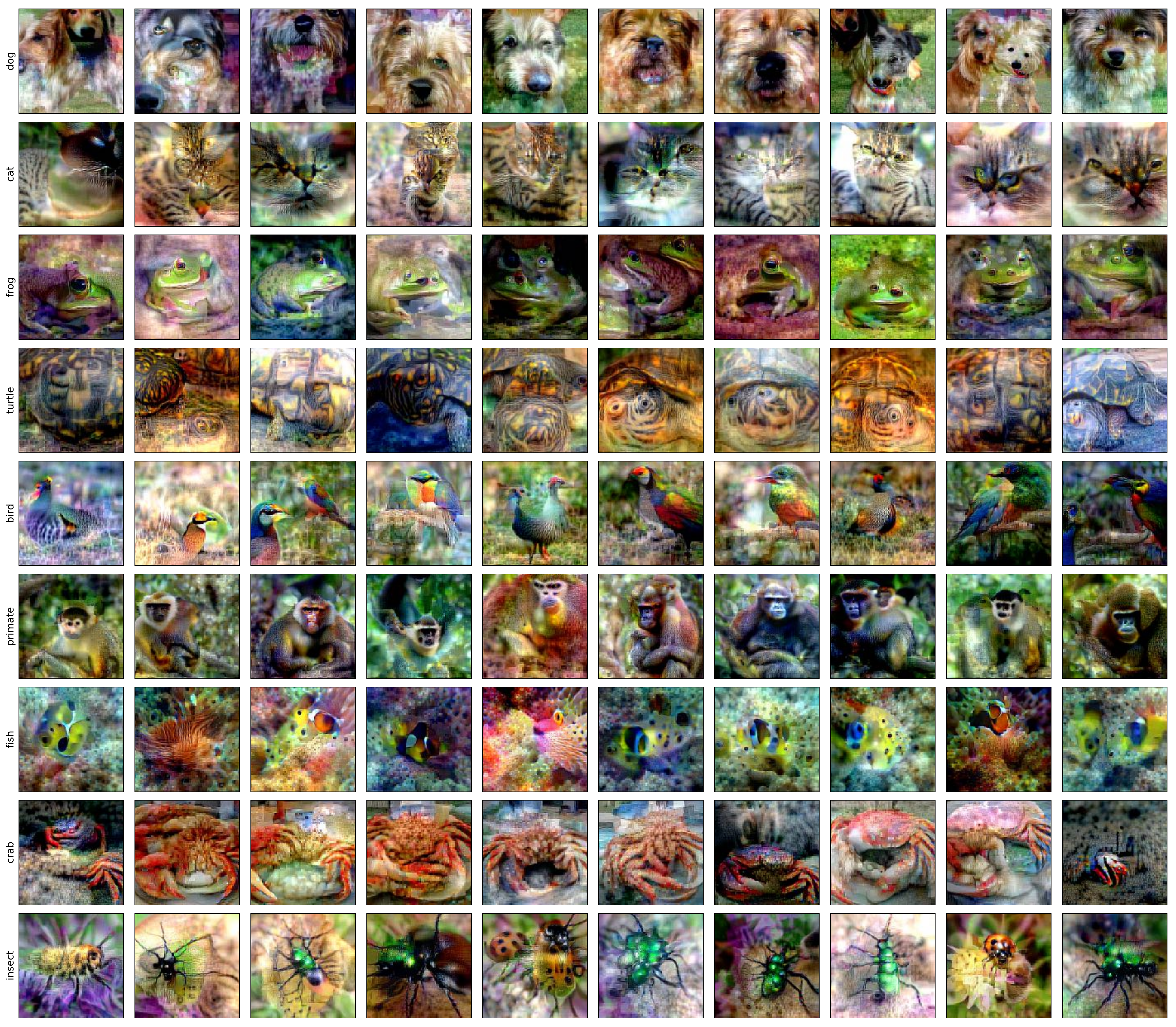}
\end{center}
\caption{Random samples generated for the Restricted ImageNet dataset.}
\label{fig:restricted_gen_full}
\end{figure}

\begin{figure}[!htp]
\begin{center}
    \includegraphics[width=\textwidth]{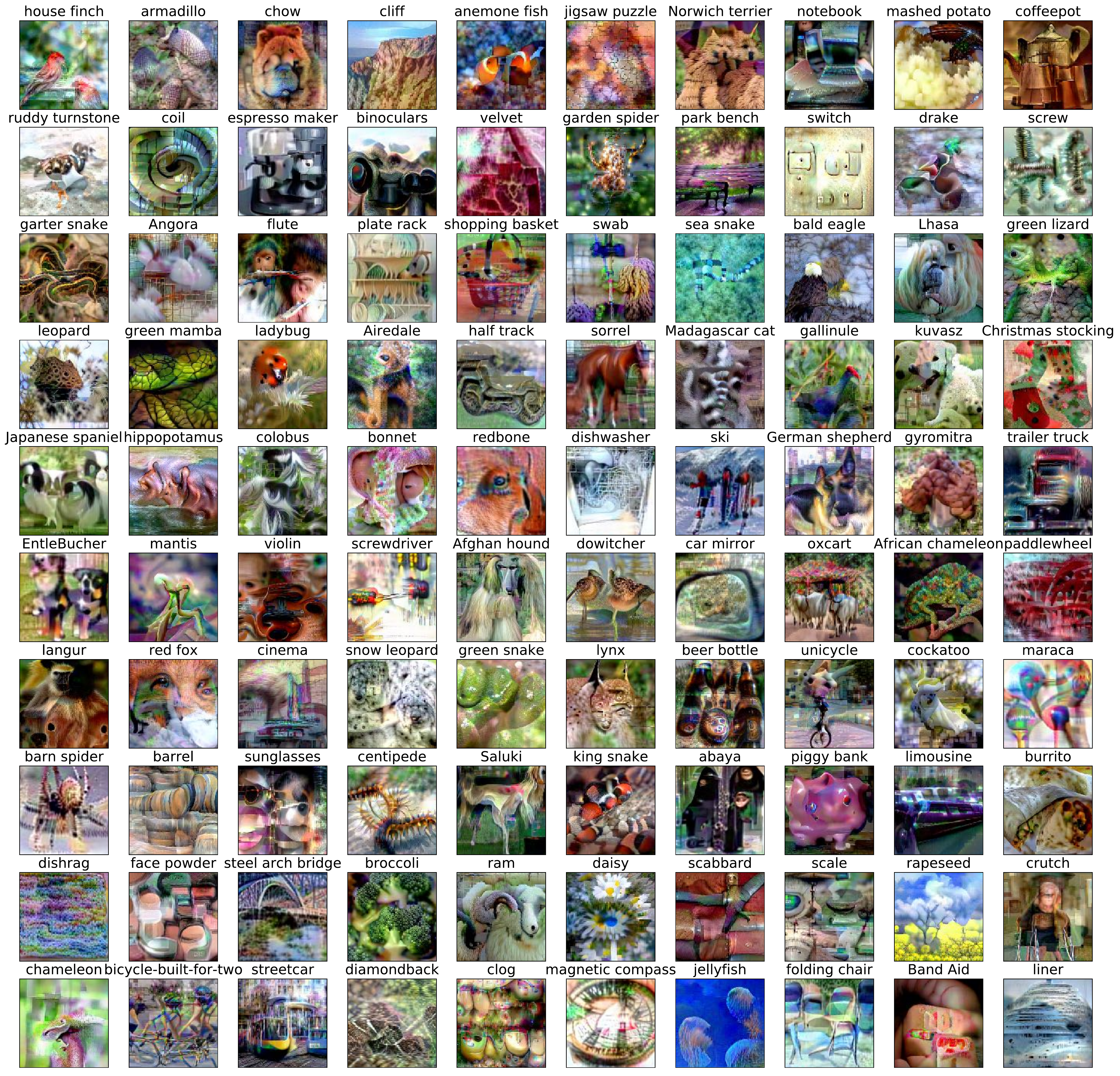}
\end{center}
\caption{Random samples generated for the ImageNet dataset.}
\label{fig:imagenet_gen_full}
\end{figure}

\begin{figure}[!htp]
\begin{center}
    \includegraphics[width=\textwidth]{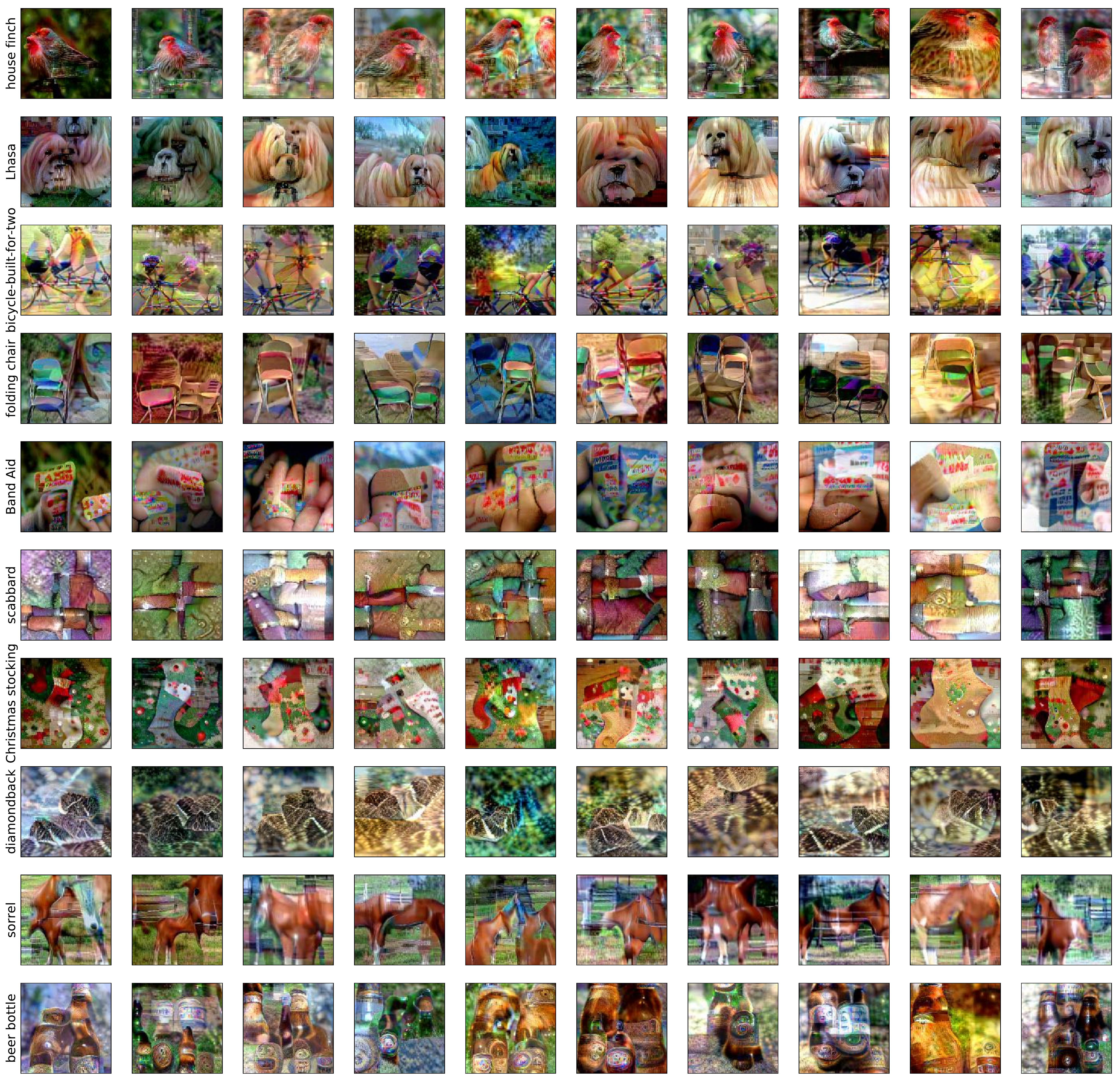}
\end{center}
\caption{Random samples from a random class subset.}
\label{fig:diversity}
\end{figure}

\begin{figure}[!htp]
\begin{center}
    CIFAR10\\[.15cm]
    \includegraphics[width=\textwidth]{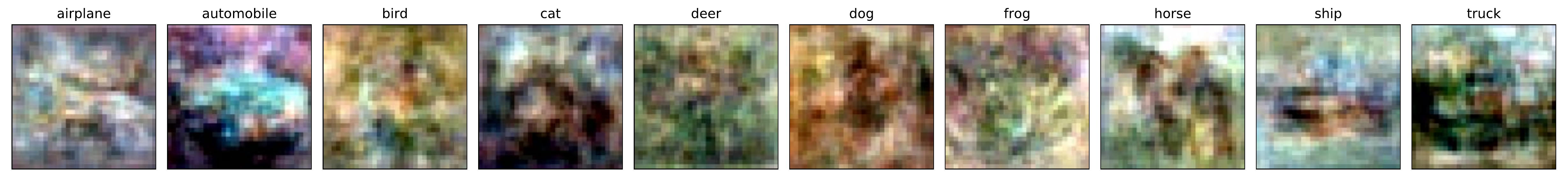}\\
    Restricted ImageNet \\[.15cm]
    \includegraphics[width=\textwidth]{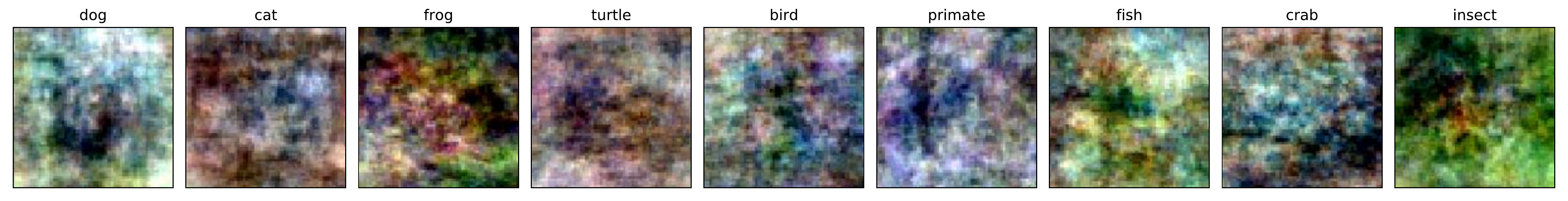}\\
    ImageNet\\[.15cm]
    \includegraphics[width=\textwidth]{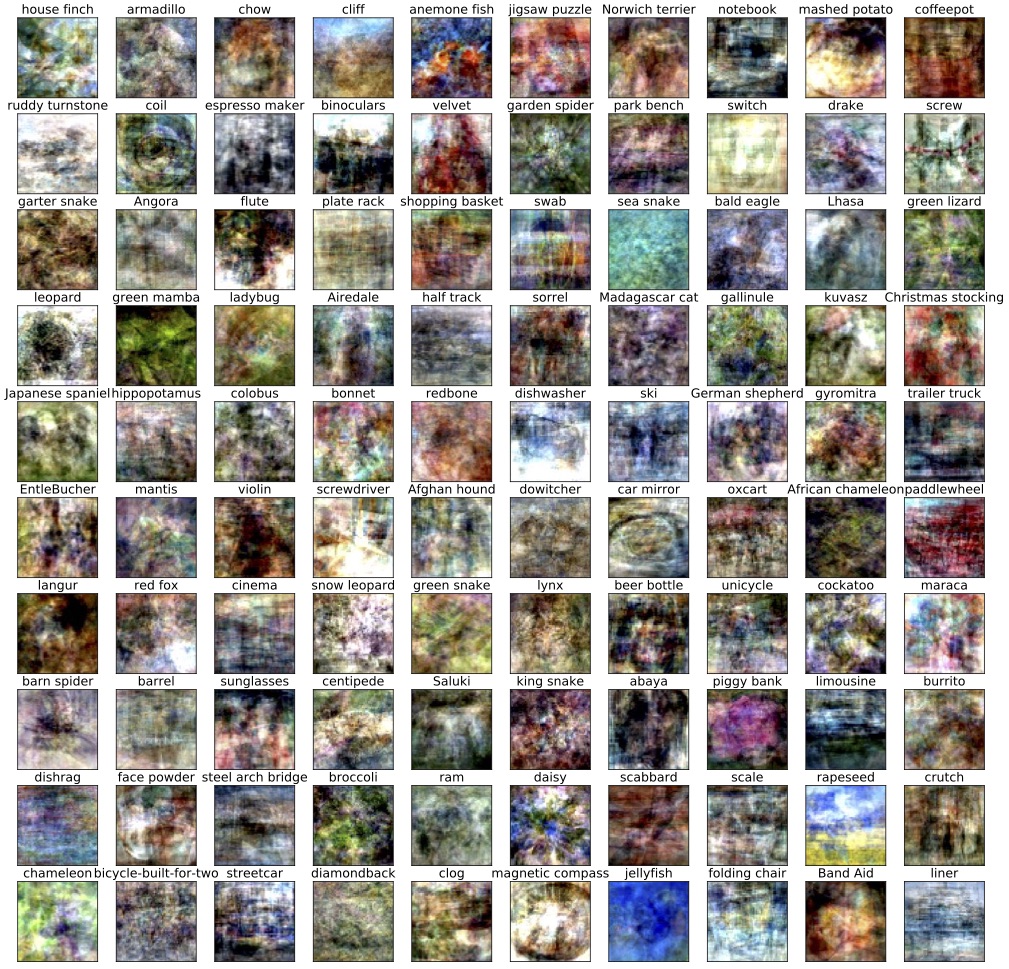}
\end{center}
\caption{Samples from class-conditional multivariate normal distributions used
as a seed for the generation process.}
\label{fig:seeds}
\end{figure}